\documentclass[lettersize,journal]{IEEEtran}
\usepackage{amsmath,amsfonts}
\usepackage{algorithmic}
\usepackage{algorithm}
\usepackage{array}
\usepackage[caption=false,font=normalsize,labelfont=sf,textfont=sf]{subfig}
\usepackage{textcomp}
\usepackage{stfloats}
\usepackage{url}
\usepackage{verbatim}
\usepackage{graphicx}
\usepackage{cite}
\usepackage{hyperref}
\usepackage{cleveref}
\usepackage{subfig}
\usepackage{multirow}
\usepackage{amsthm}
\usepackage{xcolor}
\begin{document}

\title{Sustaining Plasticity via Learnable Wavelet Activations in Continual Learning}

\author{Zeyang Zhang, Tieliang Gong$^*$, Junyan Lu, Weizhan Zhang
\thanks{$^*$Corresponding author: Tieliang Gong.}

\thanks{Z. Zhang (zzy3123351021@stu.xjtu.edu.cn), T. Gong (adidasgtl@gmail.com), J. Lu (lujunyan@stu.xjtu.edu.cn) and W. Zhang (zhangwzh@mail.xjtu.edu.cn) are with the School of Computer Science and Technology, Xi’an Jiaotong University and Institute of Multimedia Knowledge Fusion and Engineering, Xi’an 710049, China.}
}

\markboth{Journal of \LaTeX\ Class Files,~Vol.~14, No.~8, August~2021}%
{Shell \MakeLowercase{\textit{et al.}}: A Sample Article Using IEEEtran.cls for IEEE Journals}

\IEEEpubid{0000--0000/00\$00.00~\copyright~2021 IEEE}

\maketitle

\begin{abstract}
Plasticity loss has emerged as a critical challenge in continual learning that significantly hinders the acquisition of sequential tasks. While optimizing activation designs offers a potential solution, current fixed-form functions suffer from an inherent spectral bias towards low-frequency variations, whereas learnable variants permit unconstrained updates that induce catastrophic forgetting. To address these limitations, we propose a novel learnable wavelet activation that decomposes the activation function into low-frequency and high-frequency components to explicitly counter spectral bias. Furthermore, we employ dynamic wavelet injection to adaptively enhance plasticity for new tasks, alongside a regularization strategy to ensure the stability of previous learned knowledge. Theoretically, we provide rigorous mathematical guarantees for the proposed framework, proving the structural necessity of the hybrid wavelet architecture for efficient $L^2$ approximation and demonstrating that the decoupled learning rate mechanism successfully restores network plasticity for high-frequency information. Additionally, we provide a formal derivation of the loss-driven injection trigger mechanism to precisely guide the injection. Extensive empirical evaluations demonstrate that our approach maintains superior trainability and generalization throughout the learning process and achieves state-of-the-art performance across diverse continual learning benchmarks. 
\end{abstract}


\section{Introduction}
Continual learning requires neural networks to acquire new knowledge from a sequence of tasks over time without erasing previously learned information \cite{rebuffi2017icarl}. This paradigm imposes a fundamental conflict between the competing requirements of plasticity for adapting to new data and stability for preserving prior knowledge \cite{lu2025rethinking}. While extensive research has addressed catastrophic forgetting through replay buffers \cite{tong2025coreset,gu2024summarizing}, regularization constraints \cite{hassan2024new,kirkpatrick2017overcoming} and parameter isolations \cite{zhou2022model,jin2023growing}, recent studies have identified a critical issue known as plasticity loss, where models progressively lose the capacity to adapt to new tasks even when catastrophic forgetting is mitigated \cite{xu2023drm,lewandowski2023directions,lee2024slow,nikishin2023deep}.

\begin{figure}[!t]
\centering
\includegraphics[width=\columnwidth]{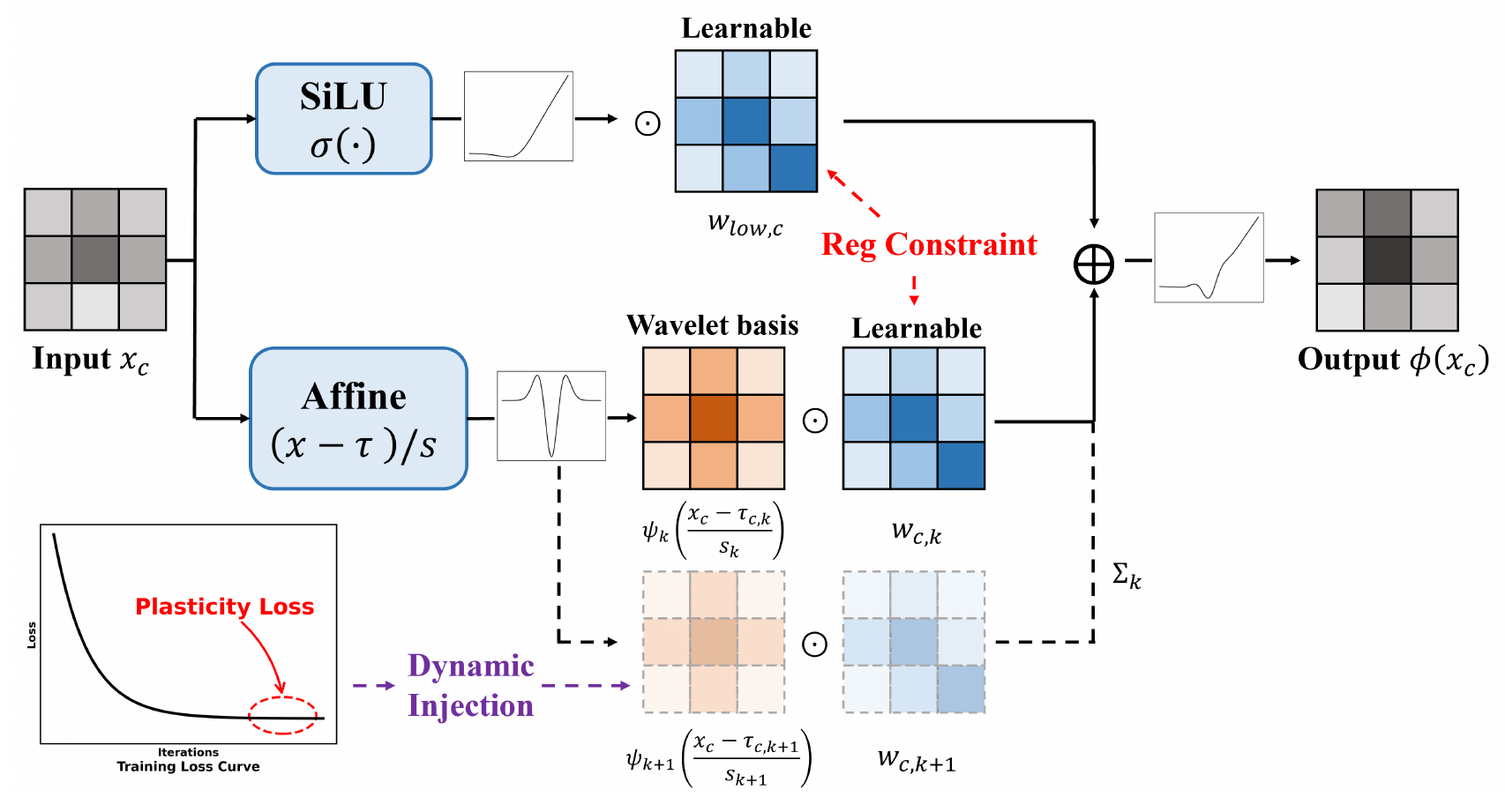}
\caption{Schematic illustration of the ChannelWavAct framework.}
\label{fig_1}
\end{figure}

Current explanations for plasticity loss span multiple levels such as dead neurons \cite{sokar2023dormant}, large weight norms \cite{lyle2023understanding}, feature rank \cite{kumar2020implicit}, the sharpness of the loss landscape \cite{lyle2023understanding} and noise memorization \cite{shin2024dash}. Existing works have primarily focused on resetting inactive units \cite{dohare2024loss,elsayed2024addressing}, regularizing weights \cite{lewandowski2024learning,elsayed2024weight}, optimizing feature spaces \cite{he2024adaptive,lyle2022understanding} and designing novel architectures \cite{nauman2024bigger,obando2024value}. While existing approaches primarily address plasticity loss through optimization constraints or architectural modifications, they frequently neglect the fundamental role of the activation function.

In fact, standard activation functions exhibit a structural impediment to continual learning.  Although recent approaches like AID and Randomized Smooth-Leaky \cite{park2025activation,lillo2025activation} introduce stochastic mechanisms to sustain plasticity, these methods, similar to the standard ReLU \cite{maas2013rectifier}, remain bound by fixed functional forms. Consequently, these activations exhibit an inherent spectral bias towards low-frequency functions, prioritizing the learning of global variations while struggling to capture the local details essential for complex tasks \cite{rahaman2019spectral,morsali2025staf}. Conversely, fully parameterized paradigms such as Rational approximations and KAN \cite{delfosse2021adaptive,liu2024kan} leverage learnable basis functions to enable arbitrary functional adaptation, yet they lack explicit constraints preserving prior task structure. During optimization on new task , gradient descent freely modifies activation parameters to minimize current loss, which can perturb previously learned parameters configurations. This unconstrained plasticity reintroduces catastrophic forgetting at the activation level, independent of weight-space stability mechanisms \cite{rebuffi2017icarl}.

\IEEEpubidadjcol

To address aforementioned challenges, we propose a novel wavelet-based activation framework that synergizes local adaptability with global stability. Our approach formulates the activation function as a learnable composition of a global base approximation and local wavelet details, incorporating a dynamic wavelet injection mechanism to actively introduce high-frequency components. This design explicitly counters the spectral bias of static functions, thereby enhancing the network's plasticity. Simultaneously, to mitigate the catastrophic forgetting common in learnable approaches, we introduce a targeted parameter regularization strategy. By imposing constraints on the activation slopes, it maintains the stability of previously learned functional configurations. Comprehensive experiments across multiple continual learning benchmarks demonstrate that our method effectively alleviates plasticity loss and achieves superior performance compared to state-of-the-art baselines.

\subsection{Related Work}

Continual Learning (CL) envisions models that accumulate knowledge sequentially from dynamic data streams while preventing the erasure of previously consolidated information \cite{rebuffi2017icarl}. The central challenge in this paradigm lies in reconciling the stability-plasticity dilemma \cite{lu2025rethinking}. Traditionally, the literature has prioritized the stability aspect, with methodologies predominantly categorized into replay-based \cite{tong2025coreset,gu2024summarizing}, regularization-based \cite{hassan2024new,kirkpatrick2017overcoming} and parameter isolation-based methods \cite{zhou2022model,jin2023growing}. Recently, a growing body of literature has identified the loss of model plasticity as a critical bottleneck in long-sequence learning \cite{dohare2021continual,lyle2023understanding}. To mitigate this degradation, current strategies have primarily focused on macroscopic interventions, such as weight re-initialization \cite{dohare2024loss,elsayed2024addressing}, parameter Regularization \cite{lewandowski2024learning,elsayed2024weight} and feature space optimization \cite{he2024adaptive,lyle2022understanding}. Distinct from these approaches, we address the plasticity loss directly through the lens of activation functions.

Activation functions serve as the primary gatekeepers of gradient information, determining how much learning signal survives backpropagation. While the dominant ReLU \cite{maas2013rectifier} suffers from the irreversible dead-unit problem where neurons become permanently inactive \cite{sokar2023dormant}, recent approaches have introduced stochastic interventions to mitigate this issue. Specifically, AID \cite{park2025activation} employs an interval-wise dropout mechanism to regularize the network towards pseudo-linear behavior, while Randomized Smooth-Leaky \cite{lillo2025activation} integrates function smoothness with randomized negative slopes to sustain robust gradient propagation. However, these methods essentially remain variants of static activation functions. Restricted to pre-defined functional forms, they cannot adaptively evolve to capture the evolving high-frequency details necessary for continual learning. More recently, KAN \cite{liu2024kan} and Adaptive Rational Activations \cite{delfosse2021adaptive} have introduced learnable B-splines and Padé approximants to achieve superior approximation. However, these approaches lack mechanisms to safeguard prior knowledge, where unrestricted adaptation inevitably leads to catastrophic forgetting. We address these limitations by proposing a wavelet activation framework that incorporates dynamic wavelet injection to ensure plasticity and parameter regularization to strictly safeguard historical stability.

\subsection{Main Contributions}

Our main contributions in this work are as follows:
\begin{itemize}
    \item \textbf{Wavelet-based activation framework:} We introduce ChannelWavAct, which decomposes activations into a fixed low-frequency base for stability and learnable high-frequency wavelets for plasticity. This explicitly achieves adaptability to local information while capturing global information. 
    \item \textbf{Adaptive capacity expansion with stability preservation:} We propose (i) dynamic wavelet injection triggered by loss stagnation to expand model capacity when plasticity saturates, and (ii) slope-specific regularization that selectively penalizes wavelet magnitude parameters while permitting geometric parameters to adapt to distribution shifts.
    \item \textbf{Rigorous theoretical foundation:} We prove the necessity of our decoupled learning rate mechanism, loss-stagnation trigger, and hybrid wavelet architecture. Furthermore, we provide formal guarantees for the $L^2$ approximation capability of ChannelWavAct and its effectiveness in restoring model plasticity for high-frequency components.
    \item \textbf{Superior performance:} Extensive experiments validate that our proposed wavelet activation exhibits optimal plasticity retention. Our approach consistently outperforms existing baseline activation functions and achieves state-of-the-art performance across a diverse range of continual learning benchmarks.
\end{itemize}

\section{Preliminary}

\subsection{Problem Setting}

We consider the standard Class-Incremental Learning (CIL) setting, where a model $f_{\theta}$ learns from a sequence of $T$ tasks, denoted as $\mathcal{S} = \{\mathcal{T}_1, \mathcal{T}_2, \dots, \mathcal{T}_T\}$. Each task $\mathcal{T}_t$ contains a dataset $\mathcal{D}_t = \{(x_i, y_i)\}_{i=1}^{N_t}$, where $x_i \in \mathcal{X}_t$ is the input image and $y_i \in \mathcal{Y}_t$ is the corresponding label. The label spaces between tasks are disjoint, i.e., $\mathcal{Y}_t \cap \mathcal{Y}_{t'} = \emptyset$ for $t \neq t'$. During the training of task $\mathcal{T}_t$, the model only has access to $\mathcal{D}_t$. The objective is to minimize the empirical risk over all seen tasks up to the current time step $t$:
\begin{equation}
    \mathcal{L}_{total}(\theta) = \mathbb{E}_{(x,y) \sim \mathcal{D}_t} [\ell(f_\theta(x), y)]
\end{equation}
where $\ell(\cdot)$ is the cross-entropy loss. For simplicity, we decompose the model parameters $\theta$ into weight parameters $\mathbf{W}$ and bias parameters $\mathbf{b}$.

\subsection{Kolmogorov-Arnold Networks (KAN)}

The Kolmogorov-Arnold Network (KAN) has recently emerged as a promising alternative to Multi-Layer Perceptrons (MLP). While MLP are theoretically grounded in the Universal Approximation Theorem (UAT), KAN draw inspiration from the Kolmogorov-Arnold Representation Theorem (KAT) \cite{lin1993realization}.The KAT asserts that any multivariate continuous function $f(\mathbf{x})$ defined on a bounded domain can be represented as a finite composition of continuous univariate functions and the binary operation of addition. Formally, the theorem is expressed as follows:
\begin{equation}
    f(\mathbf{x}) = f(x_1, \dots, x_n) = \sum_{q=0}^{2n} \Phi_q \left( \sum_{p=1}^n \phi_{q,p}(x_p) \right)
\end{equation}
where $\Phi_q$ and $\phi_{q,p}$ are univariate functions for each variable. KAN parameterize these univariate functions $\phi$ using a linear combination of a basis function $b(x) = \text{silu}(x) = x / (1 + e^{-x})$ and a B-spline function $\text{spline}(x) = \sum_i c_i B_i(x)$. The resulting learnable activation function $\phi(x)$ is defined as:
\begin{equation}
    \phi(x) = w_b \cdot b(x) + w_s \cdot \text{spline}(x)
\end{equation}
where the $w_b$ and $w_s$ represent learnable parameters that control the overall magnitude of the activation function. Consequently, a KAN layer can be expressed as:
\begin{equation}
    x_{l+1} = \underbrace{
\begin{pmatrix}
\phi_{l,1,1}(\cdot) & \cdots & \phi_{l,1,n_l}(\cdot) \\
\phi_{l,2,1}(\cdot) & \cdots & \phi_{l,2,n_l}(\cdot) \\
\vdots & \ddots & \vdots \\
\phi_{l,n_{l+1},1}(\cdot) & \cdots & \phi_{l,n_{l+1},n_l}(\cdot)
\end{pmatrix}
}_{\Phi_l} x_l
\end{equation}
where $\mathbf{x}_l$ and $\mathbf{x}_{l+1}$ denote the input and output vectors of the layer, respectively, and $\mathbf{\Phi}_l$ represents the matrix of 1D univariate functions for layer $l$. A complete KAN network is constructed by stacking multiple such KAN layers.

\section{Learnable Wavelet-based Activation}

Prevalent continual learning architectures rely on fixed-form activation functions, such as the Rectified Linear Unit (ReLU). Despite their efficiency, these functions often suffer from the dead-unit problem: neurons that output 0 receive no gradient and thus become inactive permanently. This permanent loss of capacity progressively diminishes the model's plasticity, severely constraining its adaptability to subsequent tasks.

\begin{figure}[!t]
\centering
\includegraphics[width=\columnwidth]{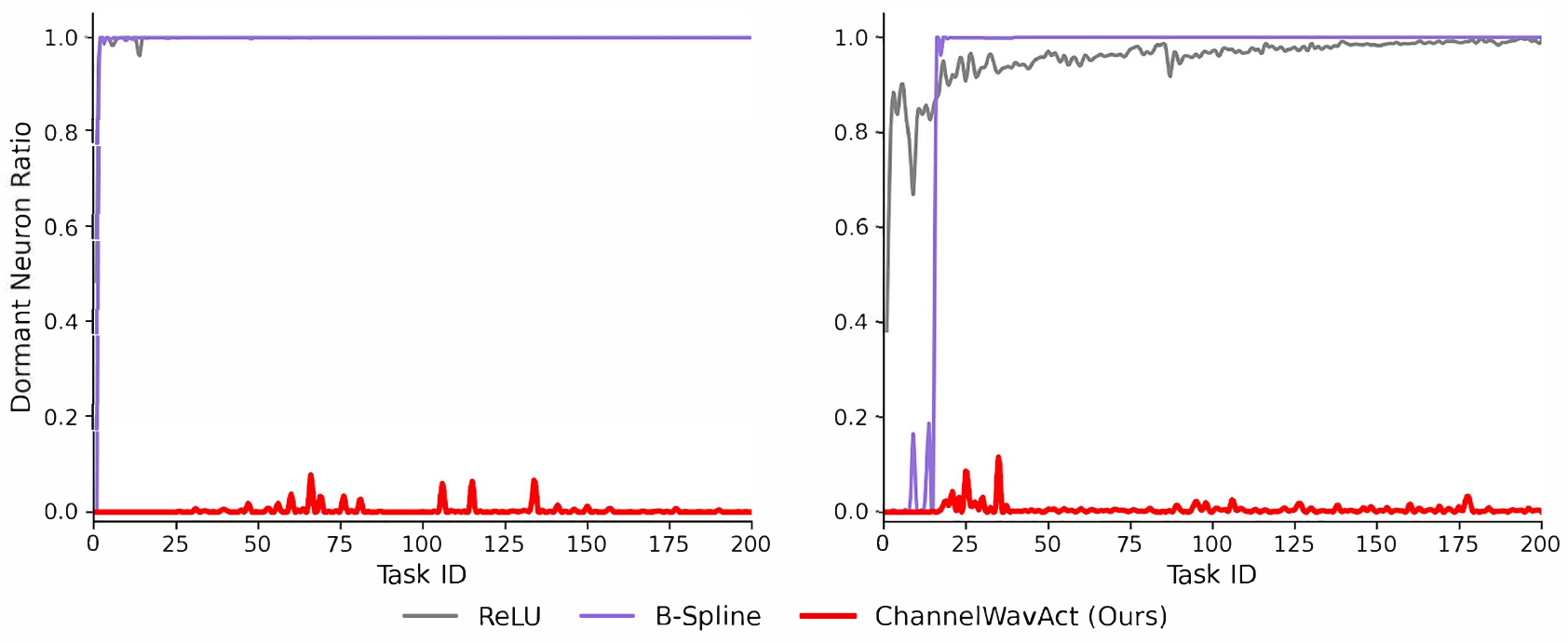}
\caption{\textbf{Evolution of the Dormant Neuron Ratio under Input and Label Trainability settings.} The left and right panels display the dormant neuron ratio across sequential tasks for Permuted MNIST and Random Label MNIST.}
\label{fig:Dormant Neuron Ratio}
\end{figure}

Kolmogorov-Arnold Networks (KAN) attempt to solve the dead-unit problem by replacing fixed nodes with learnable activation functions. However, the original KAN relies on B-splines, which works well for simple tabular data but struggles with high-dimensional features. In these deep learning contexts, spline-based KANs become inefficient and memory-intensive, making them difficult to use in practice. Specifically, as shown in \Cref{fig:Dormant Neuron Ratio}, current methods still suffer from neuron dormancy, where the number of inactive neurons rises rapidly to a saturation point.

To address these limitations, we introduce a novel wavelet-based activation. Unlike splines which are computationally heavy, our approach leverages wavelets for their efficiency. This allows us to explicitly decompose feature signals into two parts: a stable low-frequency component to preserve existing knowledge and some flexible high-frequency components to quickly adapt to new tasks. This design naturally achieves a better balance between stability and plasticity.

\subsection{Channel-wise Learnable Wavelet Activation}

To overcome the scalability limitations of KAN while keeping the benefits of learnable activations, we propose the Channel-wise Wavelet Activation (ChannelWavAct). This activation is inspired by the Discrete Wavelet Transform (DWT), which builds signals by combining a stable low-frequency base with high-frequency details \cite{bozorgasl2405wav}. Designed to replace standard activations in ResNet backbones, ChannelWavAct operates independently on each channel, learning a task-specific non-linear transformation by combining a base activation with a set of wavelet bases.

Let $x_{c}$ denote the input feature map for channel $c$. We define the activation function $\Phi_c(x_c)$ as:
\begin{equation}
    \Phi_c(x_c) = \underbrace{w_{low, c} \cdot \sigma(x_c)}_{\text{Approximation (Low-Freq)}} + \underbrace{\sum_{k=1}^{K} w_{c,k} \cdot \psi\left(\frac{x_c - \tau_{c,k}}{s_k}\right)}_{\text{Detail (High-Freq)}} \label{eq 4}
\end{equation}
Here, $w_{c,k}$ controls the amplitude of the $k$-th wavelet in channel $c$. The parameter $\tau_{c,k}$ shifts the wavelet's position, allowing the network to focus on specific parts of the input. $s_k$ denotes the scale that controls the bandwidth (frequency) of the wavelet. In our implementation, we define this as $\exp(\log s_k)$ to ensure the value stays positive.

Our formulation mimics the inverse DWT by combining a base component for approximation with a wavelet component for details. For the base, we use $\sigma(x)=SiLU(x)$ to capture global low-frequency trends to ensure stability. For the wavelet term, we use the Mexican Hat wavelet $\psi(u) = (u^2 - 1)\exp(-u^2/2)$ to handle local high-frequency variations. By optimizing the learnable weights $w_{c,k}$, shifts $\tau_{c,k}$, and shared scales $s_k$, the network dynamically refines its feature extraction. We share $s_k$ to keep the frequency structure consistent, allowing $w_{c,k}$ and $\tau_{c,k}$ to vary by channel. This flexibility ensure the model to capture complex features that fixed-form functions simply miss.

\subsection{Dynamic Wavelet Injection and Stability Regularization}
\label{sec4.2}

While replacing static ReLUs with learnable activations enhances the model's immediate representational power, utilizing a fixed configuration of learnable parameters remains insufficient for continual learning. Most current methods typically constrain the model to a static parameter space. In class-incremental settings, this rigidity creates a dilemma: optimizing the shared activation parameters to accommodate new task distributions inherently changes the feature response for previously learned tasks, causing catastrophic forgetting. Conversely, restricting parameter updates to preserve stability imposes a hard capacity limit, leading to plasticity exhaustion as the sequence of new tasks progresses.

To effectively navigate the stability-plasticity dilemma, we propose dynamic wavelet injection to actively expand model capacity when plasticity is exhausted and stability regularization to strictly preserve knowledge encoded in previous tasks.

\subsubsection{Dynamic Wavelet Injection}

We propose a dynamic mechanism that actively monitors the training dynamics and injects additional learnable wavelet bases when plasticity saturation is detected.

\textbf{Loss-Based Trigger Mechanism.} We introduce a monitoring strategy to detect plasticity loss. Let $\mathcal{L}^*_{min}$ be the minimum training loss observed so far within the current epoch. At iteration $i$, given the current batch loss $\mathcal{L}_i$ and a relative margin $\delta$, we formalize the update rule for the stagnation counter $C_{bad}^{(i)}$ and the injection trigger condition as follows:
\begin{equation}
    C_{bad}^{(i)} = \begin{cases} C_{bad}^{(i-1)} + 1 & \text{if } \mathcal{L}_i \geq \mathcal{L}^*_{min} \cdot (1 - \delta) \\ 0 & \text{otherwise} \end{cases}
\end{equation}
Here, the counter $C_{bad}$ accumulates when the loss fails to show a sufficient drop and resets whenever a new optimum is found. The injection procedure is initiated strictly when this counter exceeds the pre-defined patience threshold $P$ (i.e., $C_{bad} > P$), indicating that the representational capacity of the current activation basis is saturated.

\textbf{Wavelet Basis Expansion.} Once triggered, we expand the number of wavelets in the activation function for each channel $c$. Let $K_{old}$ be the current number of bases. We inject $\Delta$ additional wavelet basis functions, updating the total capacity to $K_{new} = K_{old} + \Delta$. To ensure these newly added components effectively capture the fine-grained features that the current model fails to resolve, we employ a targeted initialization strategy focused on high-frequency details. Specifically, the parameters for the new bases are set as follows: 
\begin{itemize}
    \item \textbf{Scale Initialization $s_k$:} To ensure positive values and a smooth frequency distribution, we optimize $s_k$ in the logarithmic space. When adding new wavelets, we initialize their scales based on the average of the current ones, i.e. $s_{new} \leftarrow \max(\epsilon, \frac{1}{K}\sum s_{i})$. This approach prevents abrupt shifts in frequency.
    \item \textbf{Translation $\tau_{c,k}$ and Weight $w_{c,k}$:} We adopt a zero-initialization strategy where both the translation and weight parameters for the new wavelets are set to zero. This ensures new components initially contribute zero output, preventing sudden feature space perturbations and allowing the model to integrate high-frequency details gradually during training.
\end{itemize}

\subsubsection{Slope-Specific Regularization}

While injection provides plasticity, we must prevent the modification of existing bases from erasing old knowledge. We employ a partial regularization strategy that acts solely on the old coefficients, decoupling stability from plasticity.

When transitioning from task $t$ to $t+1$, we save a snapshot of the current activation parameters: $\theta_{old} = \{ w_{1:K}, \tau_{1:K}, s_{1:K} \}$. During the training of task $t+1$ (where the total number of bases increases to $K_{total}$), we add a regularization term to the objective function:
\begin{equation}
    \mathcal{L}(\theta) = \mathcal{L}_{CE} + \lambda \mathcal{L}_{reg}
\end{equation}
\begin{equation}
    \begin{split}
        \mathcal{L}_{reg}(\theta) = \sum_{c=1}^{C} \bigg( & \lambda_{low} \| w_{low,c} - w_{low,c}^{old} \|^2 \\ & + \lambda_{wav} \sum_{k=1}^{K_{old}} \| w_{c,k} - w_{c,k}^{old} \|^2 \bigg)
    \end{split}
\end{equation}
where $K_{old}$ represents the number of bases before the current injection and the summation is limited to the range $k=1 \dots K_{old}$, covering only the components that existed before the expansion.

Here the newly injected bases ($k=K_{old}+1, \dots, K_{old} + \Delta$) are not subject to regularization. This allows the new units to fully adapt to the new task. We choose to selectively penalize the weights ($w_{low}$ and $w_{c,k}$) while leaving the geometric parameters (translations $\tau$ and scales $s$) free. The weights control the strength of feature signals. Thus constraining these parameters preserves the core features learned in previous tasks, effectively maintaining memory. Conversely, $\tau$ and $s$ determine the position and frequency of the activation. By allowing these geometric parameters to adjust freely, we enable the existing bases to handle small shifts in the data distribution during continual learning.

\subsubsection{Decoupled Optimization and Post-Activation Normalization}

The use of learnable activation functions and dynamic parameter injection requires us to adjust standard optimization methods. We introduce a specialized optimization strategy and a post-activation normalization technique to ensure stable convergence.

\textbf{Decoupled Learning Rates and Optimizer Reset.} Since the model architecture changes during each injection phase, the previous optimizer state becomes invalid. Therefore, we rebuild the optimizer state to accommodate the new parameter space. To strictly enforce plasticity without sacrificing basic stability, we divide the model parameters into two distinct groups with specific optimization settings: 
\begin{itemize}
    \item \textbf{Plasticity Group:} We suggest that adaptation to new tasks requires significant adjustments in both feature extraction and frequency filtering. Therefore, we assign an amplified learning rate $\eta_{high} = \lambda_{lr} \cdot \eta_{base}$ to both the backbone weights $\mathbf{W}$ and the wavelet activation parameters $\theta_{act} = \{w, \tau, s\}$. This high learning rate ensures that the model can rapidly restructure its backbone weights and activation shapes to capture the novel distributions of the incoming task, effectively preventing under-fitting. 
    \item \textbf{Stability Group:} Conversely, regarding the bias parameters $\mathbf{b}$ and batch normalization affine parameters, we maintain their learning rate at the standard base level $\eta_{base}$. These parameters generally capture baseline statistics that tend to stay consistent across tasks. Preserving a lower learning rate for these components prevents the model representation from drifting uncontrollably during the aggressive adaptation of weights and activations.
\end{itemize}

\textbf{Post-Activation Batch Normalization.} Standard ResNet blocks usually follow a Conv-BN-ReLU sequence. However, our wavelet activation involves the summation of multiple potentially unbounded functions, which can lead to significant variance shifts in the feature map distribution, especially as $K$ increases. To control this variance and ensure stable training, we apply an additional post-activation batch normalization step directly inside the ChannelWavAct module. Formally, the output of the activation function is defined as:
\begin{equation}
    \mathbf{y}_{out} = \text{BN}\left( \Phi_c(\mathbf{x}_{in}) \right)
\end{equation}
where $\Phi_c$ is the wavelet transformation defined in \Cref{eq 4}.
This ensures that regardless of the number of injected wavelets or their amplitude, the signal propagates to the next convolutional layer with a normalized distribution.

\subsection{Pseudo-code of ChannelWavAct}

The forward propagation of ChannelWavAct is detailed in \Cref{alg:channelwavact_forward}. The process operates channel-wise. For each channel $c$, the activation $\Phi_c(x_c)$ aggregates a base term $\sigma(x_c)$ with a weighted sum of $K$ wavelet bases, which use learnable translation $\tau_{c,j}$ and scaling $s_j$ to capture multi-scale details. To mitigate internal covariate shifts from this aggregation, a Post-Activation Batch Normalization (BN) is applied to each channel before concatenating the outputs into the final tensor $y$.
\begin{algorithm}[tb]
   \caption{ChannelWavAct Forward Pass}
   \label{alg:channelwavact_forward}
\begin{algorithmic}[1]
   \STATE {\bfseries Input:} Input $x$, Wavelets Numbers $K$, Parameters $\theta = \{w, \tau, s\}$
   \STATE {\bfseries Output:} Activation Function $y=\Phi(x)$

   \FOR{channel $c = 1$ {\bfseries to} $C$}
       \STATE $\Phi_c(x_c) = w_{low,c}\sigma(x_c) + \sum_{j=1}^{K} w_{c,j}\psi_{j}(\frac{x_c-\tau_{c,j}}{s_j})$
       \STATE Post-Activation Batch Normalization: $y_c=BN(\Phi_c(x_c))$
   \ENDFOR
   
   \STATE $y=\Phi(x) = \text{Concatenate}(\{y_1, \dots, y_C\})$
   
   \STATE \textbf{Return} $y=\Phi(x)$
\end{algorithmic}
\end{algorithm}

\begin{algorithm}[tb]
   \caption{Training Procedure of ChannelWavAct}
   \label{alg:training_framework}
\begin{algorithmic}[1]
   \STATE {\bfseries Input:} Sequential Tasks $\mathcal{T} = \{T_1, ..., T_N\}$, Patience $P$, Threshold $\delta$, Reg Coef $\lambda$, $\lambda_{low}$ and $\lambda_{wav}$, Learning Rate Multiplier $\lambda_{lr}$, Base LR $\eta_{base}$, Number of new wavelets $\Delta$, Wavelets Numbers $K$
   \STATE {\bfseries Initialize:} Model parameters $\theta = \{\textbf{W}, \textbf{b}, w, \tau, s\}$, Optimizer $Opt$
   
   \FOR{task $k = 1$ {\bfseries to} $N$}
       \FOR{epoch $e = 1$ {\bfseries to} $E$}
           \STATE Reset $\mathcal{L}^*_{min} \leftarrow \infty, C_{bad} \leftarrow 0$
           
           \FOR{$(x, y)$ in $D_k$}
               \STATE Model output $f(x)$
               \STATE $\mathcal{L}_{CE} = \text{CrossEntropy}(f(x), y)$
               
               \STATE {\textbf{// Loss-based Trigger Mechanism}}
               \IF{$\mathcal{L}_{CE} \geq \mathcal{L}^*_{min} \cdot (1 - \delta)$}
                   \STATE $C_{bad} \leftarrow C_{bad} + 1$
               \ELSE
                   \STATE $C_{bad} \leftarrow 0$; $\mathcal{L}^*_{min} \leftarrow \mathcal{L}_{CE}$
               \ENDIF
               
               \STATE {\textbf{// Dynamic Injection \& Optimizer Rebuild}}
               \IF{$C_{bad} > P$}
                   \FOR{channel $c = 1$ {\bfseries to} $C$}
                       \STATE Add new wavelet component $\psi_{new}$ (Update $K \leftarrow K+\Delta$)
                       \STATE Initialize new weight $w, \tau, s$
                   \ENDFOR
                   \STATE {\textbf{// Decoupled Learning Rates}}
                   \STATE $Opt \leftarrow (\{\textbf{b}, \eta_{base}\}, \{\textbf{W}, w, \tau, s, \lambda_{lr} \cdot \eta_{base}\})$
                   \STATE $C_{bad} \leftarrow 0$
               \ENDIF
               
               \STATE {\textbf{// Regularization}}
               \STATE $\mathcal{L}_{reg} = \sum_{c=1}^{C} \bigg(\lambda_{low} \| w_{low,c} - w_{low,c}^{old} \|^2 + \lambda_{wav} \sum_{j=1}^{K_{old}} \| w_{c,j} - w_{c,j}^{old} \|^2 \bigg)$
               \STATE $\mathcal{L}_{total} = \mathcal{L}_{CE} + \lambda \mathcal{L}_{reg}$
               
               \STATE Update: $Opt.step(\nabla \mathcal{L}_{total})$ 
           \ENDFOR
       \ENDFOR
   \ENDFOR
\end{algorithmic}
\end{algorithm}

The comprehensive training procedure for ChannelWavAct is formalized in \Cref{alg:training_framework}. The procedure manages plasticity through dynamic monitoring. For each batch, a counter $C_{bad}$ tracks loss ($\mathcal{L}_{CE}$) plateaus where improvement falls below margin $\delta$. Once $C_{bad}$ exceeds threshold $P$, the system injects $\Delta$ new wavelets per channel. To facilitate adaptation, the optimizer is then rebuilt using a decoupled learning rate strategy: a higher rate $\lambda_{lr} \eta_{base}$ is assigned to backbone and activation parameters ($\textbf{W}, w, \tau, s$) to enhance plasticity, while biases retain the base rate $\eta_{base}$. Finally, the model is updated via a composite objective $\mathcal{L}_{total} = \mathcal{L}_{CE} + \mathcal{L}_{reg}$, where $\mathcal{L}_{reg}$ selectively penalizes deviations in pre-existing bases ($k \le K_{old}$) to mitigate forgetting without constraining newly injected components.

\section{Theoretical Results}

We now provide rigorous theoretical foundations to establish the necessity of the proposed ChannelWavAct framework.

\textbf{Theorem 1:} Let $\Theta(t)$ be the local dynamic Neural Tangent Kernel (NTK) \cite{jacot2018neural} of a finite-width network. For a projection direction $v_i \in \mathbb{R}^N$ ($\|v_i\| = 1$) representing a local task feature at frequency $i$, assume it satisfies the approximate eigensystem $\Theta(t)v_i \approx \lambda_i(t)v_i$ with the instantaneous eigenvalue $\lambda_i(t) = v_i^T \Theta(t) v_i$. Under a uniform learning rate $\eta_{base}$, the localized residual component $u_i(t) = v_i^T u(t)$ evolves according to:
\begin{equation}
    |u_i(t)| = |u_i(0)| \exp \left( -\eta_{base} \int_{0}^{t} \lambda_i(\tau) d\tau \right)
\end{equation}
\begin{proof}
Under a single learning rate $P = \eta_{base} I$, the residual evolution equation degenerates into:
\begin{equation}
    \frac{du(t)}{dt} = -\eta_{base} \Theta(t) u(t)
\end{equation}
Multiplying both sides by the projection direction $v_i^T$ from the left:
\begin{equation}
    v_i^T \frac{du(t)}{dt} = -\eta_{base} v_i^T \Theta(t) u(t)
\end{equation}
According to the assumption, $v_i$ is the eigenvector corresponding to the instantaneous eigenvalue $\lambda_i(t)$, satisfying $v_i^T \Theta(t) = \lambda_i(t) v_i^T$. Substituting this into the above equation, we obtain:
\begin{equation}
    \frac{d(v_i^T u(t))}{dt} = -\eta_{base} \lambda_i(t) (v_i^T u(t))
\end{equation}
Defining $u_i(t) = v_i^T u(t)$, the equation can be rewritten as a first-order linear ordinary differential equation with variable coefficients:
\begin{equation}
    \frac{du_i(t)}{dt} = -\eta_{base} \lambda_i(t) u_i(t)
\end{equation}
Integrating both sides from $0$ to $t$ yields the analytical solution:
\begin{equation}
    u_i(t) = u_i(0) \exp \left( -\eta_{base} \int_{0}^{t} \lambda_i(\tau) d\tau \right)
\end{equation}
\end{proof}

According to the Dynamic Frequency Principle \cite{xu2019frequency}, the instantaneous eigenvalues associated with high-frequency features remain orders of magnitude smaller than those of low frequencies, i.e., $\lambda_{\text{high}}(\tau) \ll \lambda_{\text{low}}(\tau)$. As a result, the integral $\int_{0}^{t} \lambda_{\text{high}}(\tau) d\tau$ remains negligibly small over any finite number of training epochs, yielding a residual decay factor of $\exp(\cdot) \approx 1$. It indicates that a uniform learning rate is insufficient to suppress high-frequency residuals within a practical training horizon, which in turn limits the network's plasticity in capturing complex local features. This observation motivates a decoupled learning rate strategy, whose necessity we formally stated in Theorem 2.

\textbf{Assumption 1:} For a high-frequency task feature direction $v_{\text{high}}$, it is assumed based on the Frequency Principle \cite{xu2019frequency} that features at distinctly different frequencies exhibit near-orthogonality in the functional space. Consequently, projecting the global low-frequency bases onto this high-frequency direction yields only a negligible residual response $\epsilon(t) \ll 1$, while the localized wavelet bases dominate this subspace with an instantaneous eigenvalue $\lambda_{\text{high}}(t)$. Formally:
\begin{equation}
    \begin{split}
        \Theta_{base}(t)v_{high} &= \epsilon(t)v_{high}, \quad \text{where } \epsilon(t) \ll 1 \\
        \Theta_{high}(t)v_{high} &= \lambda_{high}(t)v_{high}
    \end{split}
\end{equation}

\textbf{Theorem 2:} Let the Neural Tangent Kernel (NTK) matrix be decomposed into a linear superposition of two sub-kernels: $\Theta(t) = \Theta_{base}(t) + \Theta_{high}(t)$. Under Assumption 1, by employing a decoupled learning rate matrix $P = \text{diag}(\eta_{base}I_{base}, \eta_{high}I_{high})$, the effective network dynamics are governed by the modified Preconditioned NTK $\tilde{\Theta}(t)$, satisfying:
\begin{equation}
    \tilde{\Theta}(t) = \eta_{base}\Theta_{base}(t) + \eta_{high}\Theta_{high}(t)
\end{equation}
\begin{proof}
Substituting the preconditioned gradient flow into the residual evolution equation, the dynamical system can be rewritten as:
\begin{equation}
\begin{split}
    \frac{du(t)}{dt} &= - [ \nabla_{\theta_{base}} f^T (\eta_{base} I) \nabla_{\theta_{base}} f \\
    &\quad + \nabla_{\theta_{high}} f^T (\eta_{high} I) \nabla_{\theta_{high}} f ] u(t) \\
    &= - (\eta_{base} \Theta_{base}(t) + \eta_{high} \Theta_{high}(t)) u(t) \\
    &= - \tilde{\Theta}(t) u(t)
\end{split}
\end{equation}
where $\tilde{\Theta}(t)$ represents the effective preconditioned NTK. Applying $\tilde{\Theta}(t)$ to the high-frequency direction $v_{high}$ and invoking the approximate orthogonality in Assumption 1, we obtain:
\begin{equation}
\begin{split}
    \tilde{\Theta}(t) v_{high} &= \eta_{base} (\epsilon(t) v_{high}) + \eta_{high} (\lambda_{high}(t) v_{high}) \\
    &= (\eta_{high} \lambda_{high}(t) + \eta_{base} \epsilon(t)) v_{high}
\end{split}
\end{equation}
This confirms that $v_{high}$ remains an eigenvector of the preconditioned operator $\tilde{\Theta}(t)$ with an amplified effective eigenvalue $\tilde{\lambda}_{high}(t) = \eta_{high} \lambda_{high}(t) + \eta_{base} \epsilon(t)$. The analytical solution for the high-frequency residual $u_{high}(t)$ is given by:
\begin{equation}
    \begin{split}
        u_{high}(t) &= u_{high}(0) \\
    &\cdot \exp \left( - \int_0^t [\eta_{high} \lambda_{high}(\tau) + \eta_{base} \epsilon(\tau)] d\tau \right)
    \end{split}
\end{equation}
Given that $\epsilon(\tau) \approx 0$ due to negligible spectral leakage, the decay rate is predominantly controlled by $\eta_{high} \lambda_{high}(\tau)$. By explicitly setting $\eta_{high} = \lambda_{lr} \eta_{base}$ with $\lambda_{lr} \gg 1$, the inherent high-frequency spectral bias is mathematically compensated, thereby restoring the instantaneous fitting capability of the wavelet bases.
\end{proof}

Notably, the decomposition $\Theta(t) = \Theta_{\text{base}}(t) + \Theta_{\text{high}}(t)$ originates from the linear combination of the global base and local wavelets, which splits their gradient operators and eliminates cross-terms under a decoupled optimization framework. Furthermore, the results of Theorem 2 demonstrate that our decoupled learning rate mechanism effectively compensates for the deficiency of the original spectral components. By setting $\eta_{\text{high}} > \eta_{\text{base}}$, this mechanism accelerates the convergence rate of high-frequency components, thereby enabling synchronous convergence across the entire spectrum. Next, we prove the rationality of the loss-driven monitoring mechanism in Theorem 3 to precisely determine the optimal triggering condition for the injection.

\textbf{Theorem 3:} Let $\mathcal{D} = \{(x_i, y_i)\}_{i=1}^N$ be the training set. At time $t$, denote the function space spanned by $K$ wavelet bases as $\mathcal{H}_K = \text{span}\{\psi_{\tau_1}, \dots, \psi_{\tau_K}\}$, the empirical residual vector as $u_K(t) = f_K(X, t) - Y \in \mathbb{R}^N$ and the local dynamic NTK matrix as $\Theta_K(t) \in \mathbb{R}^{N \times N}$. Under continuous-time gradient flow dynamics, the following properties hold: 
\begin{enumerate}
    \item Optimization stagnation is strictly characterized by the residual vector entering the approximate null space of $\Theta_K(t)$. For a given tolerance $\epsilon_{\text{tol}} > 0$, this condition is equivalent to the Rayleigh quotient collapse \cite{robin2022convergence}: 
    \begin{equation}
        \frac{u_K(t)^T \Theta_K(t) u_K(t)}{\|u_K(t)\|_2^2} \le \epsilon_{tol}
    \end{equation}
    \item Let $\Omega_{\text{res}} = \text{supp}(\mathcal{F}(u_K))$ be the fourier-domain support of the residual, $\mu(\cdot)$ be the lebesgue measure and $B_\psi$ be the effective frequency bandwidth of a single wavelet. To ensure the expanded feature space fully resolves the residual, the minimum optimal incremental capacity $\Delta^*$ satisfies: 
    \begin{equation}
        \Delta^* = \left\lceil \frac{\mu(\Omega_{res})}{B_\psi} \right\rceil
    \end{equation}
\end{enumerate}
\begin{proof}
According to the gradient flow evolution $\frac{du_K(t)}{dt} = -\Theta_K(t)u_K(t)$, the time derivative of the squared residual norm $\|u_K(t)\|_2^2$ is derived as:
\begin{equation}
    \frac{d}{dt} \|u_K(t)\|_2^2 = -2u_K(t)^T \Theta_K(t) u_K(t).
\end{equation}

Optimization stagnation occurs when the rate of residual decay falls below a threshold $\epsilon_{tol}$ relative to its current magnitude, i.e., $|\frac{d}{dt} \|u_K(t)\|_2^2| \le 2\epsilon_{tol} \|u_K(t)\|_2^2$. Rearranging this inequality yields the Rayleigh Quotient bound:
\begin{equation}
    \frac{u_K(t)^T \Theta_K(t) u_K(t)}{\|u_K(t)\|_2^2} \le \epsilon_{tol},
\end{equation}
which indicates that the residual $u_K(t)$ has entered the approximate null space of the current NTK $\Theta_K(t)$.
 
Then, let $\Delta \mathcal{H} = \text{span}\{\psi_{\tau_{new,i}}\}_{i=1}^{\Delta K}$ be the incremental function space. By Parseval's Identity \cite{pfister2017discrete}, the projection error $\mathcal{E}_{proj}$ can be expressed in the frequency domain as the energy not covered by the new wavelet filters:
\begin{equation}
    \mathcal{E}_{proj} \propto \int_{\omega \in \mathbb{R}} \left| \mathcal{F}(u_K)(\omega) \cdot \Big( 1 - \sum_{i=1}^{\Delta} |\mathcal{F}(\psi_{\tau_i})(\omega)|^2 \Big) \right| d\omega.
\end{equation}

To ensure $\mathcal{E}_{proj} \rightarrow 0$, the union of the spectral supports of the injected bases must cover the support of the residual $\Omega_{res} = \text{supp}(\mathcal{F}(u_K))$. Invoking the lebesgue measure $\mu(\cdot)$ and the covering lemma in \cite{brown2009wavelet}, we have:
\begin{equation}
    \sum_{i=1}^{\Delta} \mu(\text{supp}(\mathcal{F}(\psi_{\tau_i}))) \ge \mu(\Omega_{res}) \implies \Delta  \cdot B_\psi \ge \mu(\Omega_{res}).
\end{equation}

Thus, the minimum optimal number of bases is $\Delta^* = \lceil \mu(\Omega_{res}) / B_\psi \rceil$.
\end{proof}

Theorem 3 provides a theoretical foundation for our loss detection mechanism. Since the numerator of the Rayleigh quotient governs the time derivative of the loss, i.e., $\frac{d}{dt}\mathcal{L} = -u_K^T \Theta_K u_K$, its collapse manifests as a loss stagnation plateau. This justifies our monitoring strategy, validating its rationality as an effective proxy. Furthermore, the spectral coverage bound $\Delta^*$ quantitatively analyzes the minimal number of wavelet bases required to resolve the frequency components of the remaining error. To further justify the specific functional form and $L^2$ approximation capability of these injected components, we establish the structural necessity of our hybrid wavelet architecture in Theorem 4.

\textbf{Theorem 4:} Let $K = [-M, M] \subset \mathbb{R}$ be a compact set, and let $L^2(K)$ denote the space of square-integrable target signals. We define a hybrid feature dictionary $\mathcal{D} = \{\phi(x)\} \cup \{\psi_{s,\tau}(x)\}$, where $\phi(x) = x \cdot \sigma(x)$ serves as the global low-frequency scaling function and $\psi_{s,\tau}(x) = (1 - (s^{-1}(x-\tau))^2)e^{-(s^{-1}(x-\tau))^2/2}$ is the localized high-frequency mother wavelet. For any target signal $f \in L^2(K)$ and any approximation tolerance $\epsilon > 0$, there exists a finite set of parameters $\Theta = \{w_{low}\} \cup \{w_k, s_k, \tau_k\}_{k=1}^N$ such that the hybrid operator $\Phi(x) = w_{low}\phi(x) + \sum_{k=1}^N w_k \psi_{s_k, \tau_k}(x)$ satisfies:
\begin{equation}
    \|f(x) - \Phi(x)\|_{L^2(K)} < \epsilon
\end{equation}
\begin{proof}
A pure wavelet system $g(x) = \sum w_k \psi(\frac{x-\tau_k}{s_k})$ is composed of Mexican Hat mother wavelets $\psi(x) = (1-x^2)e^{-x^2/2}$, which satisfy $\int_{-\infty}^{\infty} \psi(x)dx = 0$. Consequently, $\int_{\mathbb{R}} g(x)dx = 0$. For any target signal $f(x)$ with a non-zero mean $C = \int_K f(x)dx \neq 0$ on a compact set $K$, a pure wavelet system requires divergent scales ($s \to \infty$) to approximate the global trend, leading to parameter instability in finite-width networks.

To resolve this, we introduce the SiLU-based scaling function $\phi(x) = x\sigma(x)$. Since $\phi(x) > 0$ for $x > 0$ and decays exponentially for $x < 0$, its integral over $K=[-M, M]$, denoted as $I_\phi$, is strictly positive. We decompose the signal as $f(x) = w_{low} \phi(x) + f_{res}(x)$. By setting the optimal weight:
\begin{equation}
    w_{low}^* = \frac{\int_K f(x)dx}{I_\phi} = \frac{C}{I_\phi},
\end{equation}
we ensure that the remaining residual $f_{res}(x)$ has a strictly zero mean, i.e., $\int_K f_{res}(x)dx = 0$.

For the zero-mean residual $f_{res} \in L^2(K)$, we apply the Plancherel Theorem \cite{stein2011fourier} to map the $L^2$ norm into the frequency domain:
\begin{equation}
    \|f_{res} - g\|_{L^2}^2 = \frac{1}{2\pi} \int_{-\infty}^{\infty} |\hat{f}_{res}(\omega) - \hat{g}(\omega)|^2 d\omega.
\end{equation}
Since $\hat{f}_{res}(0) = 0$, the energy is concentrated in the non-zero frequency bands. Utilizing the Frame Truncation Theorem \cite{stephane1999wavelet}, there exists a finite set of wavelet parameters $\{w_k, s_k, \tau_k\}_{k=1}^N$ such that the high-frequency approximation error is bounded by $\epsilon$. Combining this with the low-frequency component via the triangle inequality, we conclude:
\begin{equation}
    \|f(x) - \Phi(x)\|_{L^2(K)} = \|(w_{low}^* \phi + f_{res}) - (w_{low}^* \phi + g)\|_{L^2} < \epsilon,
\end{equation}
\end{proof}

Theorem 4 establishes the indispensability of the hybrid architecture of ChannelWavAct. Specifically, the structural necessity constraint dictates that for signals with a non-zero mean ($\int_K f(x)dx \neq 0$), a system composed solely of wavelets would require divergent scales ($s \rightarrow \infty$) to approximate global trends, leading to a catastrophic collapse in efficiency for finite-width networks. By integrating the SiLU-based scaling function with localized wavelets, ChannelWavAct ensures optimal $L^2$ approximation for both global structures and local high-frequency details.

\section{Experiments}

\subsection{Continual Learning}

To comprehensively validate the superiority of our proposed method in continual learning scenarios, we benchmark our wavelet-based activation against mainstream baselines from two critical perspectives: \textit{trainability} and \textit{generalizability}. These baselines can be broadly divided into static and learnable approaches. The static baselines include the standard ReLU \cite{maas2013rectifier}, AID \cite{park2025activation} and Randomized Smooth-Leaky activations \cite{lillo2025activation}. The learnable baselines comprise Rational Activations \cite{delfosse2021adaptive} and KAN \cite{liu2024kan}.

\subsubsection{Trainability}

To assess the input and label plasticity of our wavelet-based activation against existing baselines \cite{dohare2024loss,kumar2023maintaining}, we follow the evaluation protocol in AID \cite{park2025activation} and utilize two standard benchmarks derived from the MNIST dataset \cite{lecun2002gradient}. Specifically, for input plasticity, we employ \textbf{Permuted MNIST}, which contains 10 handwritten digit classes with 60,000 training images and 10,000 testing images. We generate a sequence of 200 tasks where the $28 \times 28$ image pixels are arbitrarily shuffled with a fixed random permutation for each task \cite{dohare2024loss}. Parallelly, to evaluate label plasticity, we use \textbf{Random Label MNIST}, which utilizes a smaller subset of 1,600 training images to simulate a rapid fitting scenario[cite: 1063, 1064]. In this setting, we construct 200 tasks where the class labels of the $28 \times 28$ images are randomly shuffled for each task while the input pixels remain fixed \cite{kumar2023maintaining}.
\begin{figure}[!t]
  \centering
  
  \subfloat[Trainability experimental results.]{
    \includegraphics[width=0.45\textwidth]{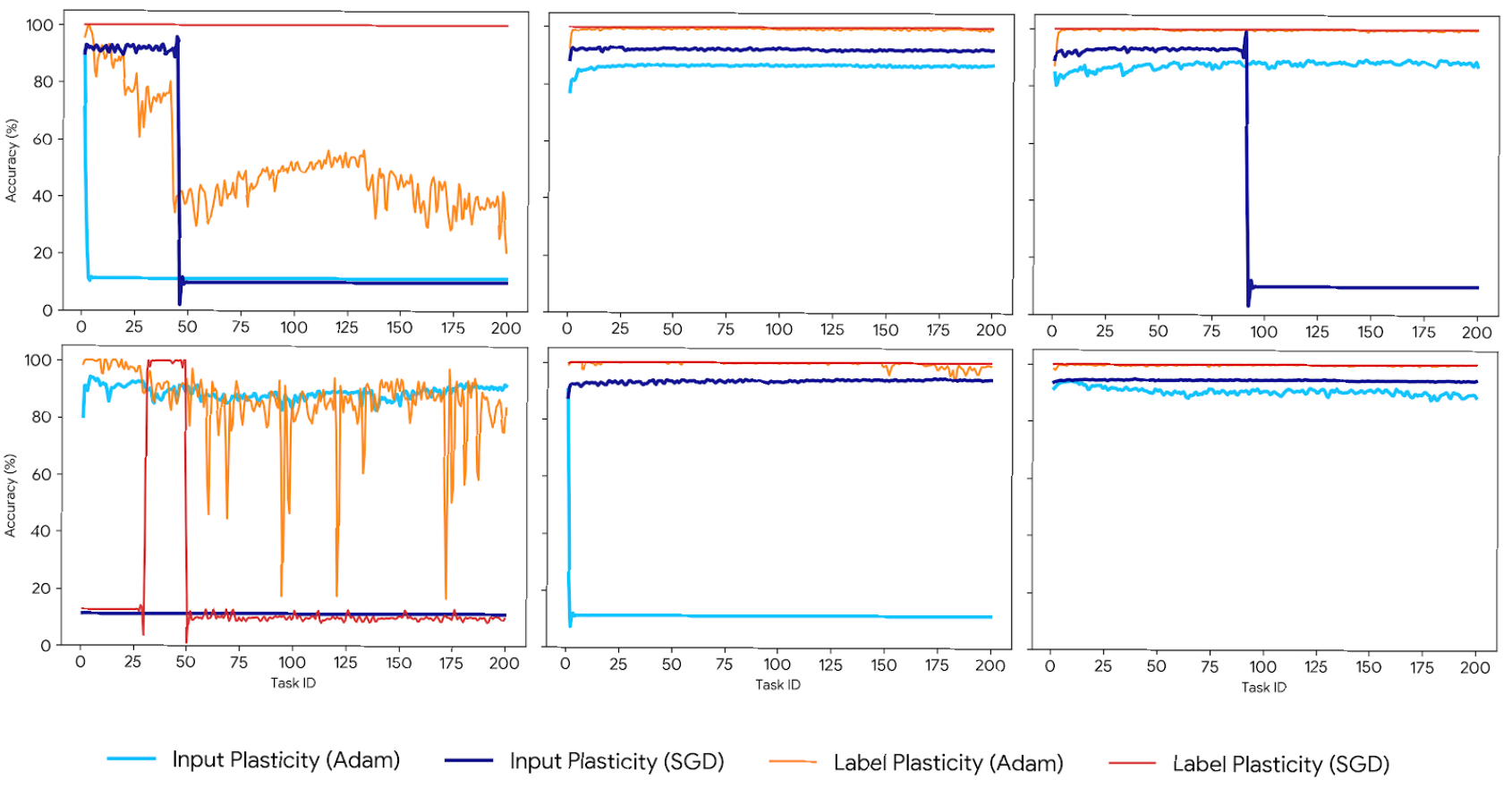}
    \label{fig:plasticity_acc}
  }
  
  \vspace{0.1in} 
  
  \subfloat[Evolution of Effective Rank (srank) across sequential tasks.]{
    \includegraphics[width=0.45\textwidth]{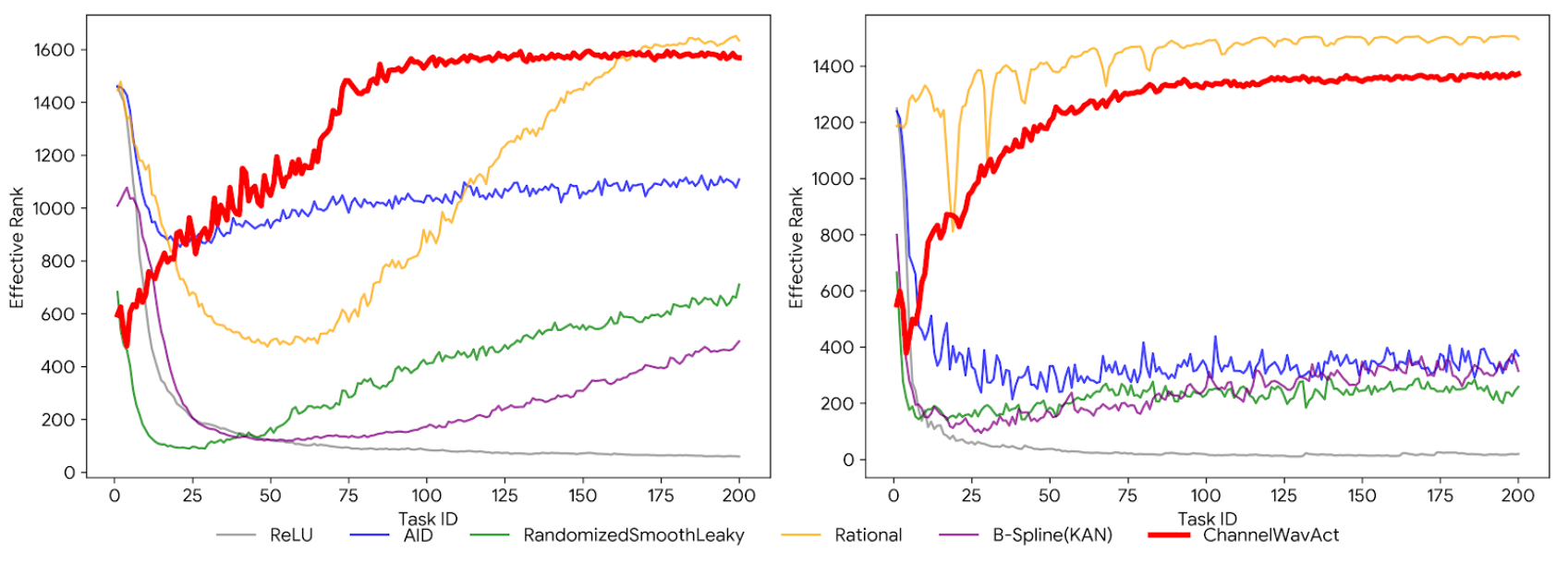}
    \label{fig:effective_rank}
  }
  
  \caption{
    (a) Accuracy of various activation functions on the Permuted MNIST and Random Label MNIST datasets. The vertical axis denotes Accuracy and the horizontal axis represents the Task ID. The plots show results for ReLU, AID, Randomized Smooth-Leaky (top row), Rational, B-spline and ChannelWavAct (bottom row). 
    (b) Evaluation of Input Plasticity on Permuted MNIST (left) and Label Plasticity on Random Label MNIST (right).
  }
  \label{fig:plasticity_result} 
\end{figure}

\Cref{fig:plasticity_acc} presents comparative results of trainability retention across continual learning of 200 tasks. First, ReLU exhibits substantial performance degradation across all tasks, particularly showing a rapid collapse in input trainability. Second, while other static baselines maintain stability, they fail to achieve optimal accuracy, particularly in input trainability scenarios, suggesting a limited trainability ceiling. Furthermore, learnable variants demonstrate inconsistent robustness: KAN fails to retain trainability in input trainability settings, exhibiting performance degradation comparable to ReLU, while Rational suffers from significant instability in label trainability tasks. In contrast, our proposed ChannelWavAct achieves superior and stable performance across all evaluated tasks, demonstrating exceptional robustness and sustained trainability.
\begin{table*}[!t]
\caption{Performance comparison on the SSD framework across Mini-ImageNet, CIFAR100 and Tiny-ImageNet datasets. All results are reported as the average of 10 independent runs.\label{tab:ssd_results}}
\centering
\setlength{\tabcolsep}{7pt}
\begin{tabular}{|l||ccc|ccc|ccc|}
\hline
\multirow{2}{*}{Method} & \multicolumn{3}{c|}{Mini-ImageNet} & \multicolumn{3}{c|}{CIFAR100} & \multicolumn{3}{c|}{Tiny-ImageNet} \\
\cline{2-10}
 & Last & Avg & Fgt & Last & Avg & Fgt & Last & Avg & Fgt \\
\hline
ReLU & 25.1$\pm$0.5 & 37.0$\pm$1.4 & 13.4$\pm$1.2 & 28.3$\pm$0.5 & 40.3$\pm$1.2 & 18.5$\pm$1.2 & 8.6$\pm$0.3 & 17.2$\pm$1.0 & 10.3$\pm$0.5 \\

AID & 21.3$\pm$0.6 & 31.1$\pm$1.4 & 8.4$\pm$0.8 & 25.5$\pm$0.6 & 35.2$\pm$1.1 & 8.8$\pm$0.9 & 7.3$\pm$0.3 & 14.1$\pm$1.0 & 6.6$\pm$0.4 \\

Randomized Smooth-Leaky & 24.1$\pm$0.3 & 36.6$\pm$0.8 & 11.7$\pm$0.7 & 28.9$\pm$0.6 & 41.7$\pm$1.0 & 15.7$\pm$0.5 & 9.1$\pm$0.2 & 18.3$\pm$0.9 & 9.5$\pm$0.7 \\

Rational & 13.3$\pm$5.3 & 18.8$\pm$6.4 & 8.6$\pm$5.2 & 17.1$\pm$3.0 & 21.6$\pm$3.4 & 7.0$\pm$2.5 & 5.7$\pm$1.6 & 10.8$\pm$2.5 & 6.1$\pm$1.5 \\

B-Spline (KAN) & 25.0$\pm$2.1 & 37.5$\pm$1.5 & 15.8$\pm$2.9 & 28.9$\pm$0.4 & 41.7$\pm$0.7 & 18.4$\pm$0.5 & 8.0$\pm$1.2 & 17.6$\pm$2.8 & 12.1$\pm$1.3 \\
\hline
ChannelWavAct & \textbf{26.2$\pm$0.4} & \textbf{39.0$\pm$0.6} & 12.7$\pm$0.5 & \textbf{30.2$\pm$0.4} & \textbf{43.5$\pm$0.9} & 16.4$\pm$0.4 & \textbf{9.9$\pm$0.2} & \textbf{19.9$\pm$0.9} & 10.9$\pm$0.4 \\
\hline
\end{tabular}
\end{table*}

\begin{table*}[!t]
\caption{Performance comparison on EWC and WA frameworks. All results were obtained with fixed random seeds and fixed task orders \cite{zhou2023pycil}.\label{tab:framework_comparison}}
\centering
\setlength{\tabcolsep}{10pt}
\begin{tabular}{|l||cccc|cccc|}
\hline
 \multirow{2}{*}{Method} & \multicolumn{4}{c|}{EWC} & \multicolumn{4}{c|}{WA} \\
\cline{2-9}
 & \multicolumn{2}{c}{CIFAR100} & \multicolumn{2}{c|}{Imagenet100} & \multicolumn{2}{c}{CIFAR100} & \multicolumn{2}{c|}{Imagenet100} \\
\cline{2-9}
 & Avg & Fgt & Avg & Fgt & Avg & Fgt & Avg & Fgt \\
\hline
\multicolumn{9}{|c|}{w/o $L_2$ Regularization} \\
\hline
ReLU & 33.4 & 56.4 & 40.0 & 71.2 & 33.0 & 82.3 & 35.8 & 83.2 \\
AID & 32.0 & 57.1 & 35.8 & 71.8 & 38.5 & 65.0 & 42.5 & 72.5 \\
Randomized Smooth-Leaky & 34.2 & 52.2 & 40.8 & \textbf{66.4} & 37.4 & 77.9 & 38.0 & 80.6 \\
Rational & 5.0 & 15.2 & 27.6 & 79.1 & 13.8 & 32.1 & 42.4 & 71.6 \\
B-Spline & 34.3 & 36.7 & $\backslash$ & $\backslash$ & 45.6 & 61.7 & $\backslash$ & $\backslash$ \\
\hline
ChannelWavAct* & 36.5 & 52.3 & \textbf{41.0} & 68.6 & 46.4 & 65.7 & 54.6 & 62.9 \\
ChannelWavAct & \textbf{41.2} & \textbf{45.3} & 40.9 & 68.4 & \textbf{55.6} & \textbf{48.2} & \textbf{64.5} & \textbf{40.8} \\
\hline
\multicolumn{9}{|c|}{$+$ $L_2$ Regularization} \\
\hline
ReLU & 31.1 & 75.9 & 28.6 & 87.8 & 63.4 & 35.1 & 67.2 & 35.8 \\
AID & 31.2 & \textbf{64.4} & 30.0 & 86.7 & 52.9 & 43.5 & 65.3 & 38.2 \\
Randomized Smooth-Leaky & 30.5 & 73.4 & \textbf{31.5} & \textbf{81.6} & 62.8 & 38.4 & 67.7 & \textbf{31.0} \\
Rational & 4.4 & 14.8 & 23.1 & 85.0 & 13.8 & 32.1 & 42.4 & 71.6 \\
B-Spline & 30.6 & 76.2 & $\backslash$ & $\backslash$ & 61.5 & 29.6 & $\backslash$ & $\backslash$ \\
\hline
ChannelWavAct w/o decoupled lr & 29.6 & 73.0 & 28.6 & 88.1 & 62.5 & 30.1 & \textbf{68.2} & 33.9 \\
ChannelWavAct & \textbf{32.5} & 70.3 & 28.0 & 86.3 & \textbf{65.7} & \textbf{29.9} & 63.9 & 34.9 \\
\hline
\end{tabular}
\end{table*}

The evolution of the Effective Rank is illustrated in \Cref{fig:effective_rank}. We strictly follow the srank definition proposed by \cite{kumar2020implicit}. It can be observed that the standard ReLU activation suffers from a precipitous decline in effective rank as tasks progress, implying that the representation space degenerates into a low-dimensional subspace that severely limits the network's capacity to encode new information. While the Rational activation maintains a high srank, its suboptimal downstream performance in \Cref{tab:ssd_results,tab:framework_comparison} suggests that these high-dimensional representations may lack semantic discriminability or contain excessive noise. In contrast, our proposed ChannelWavAct demonstrates a robust and sustained high effective rank that correlates with superior accuracy. This distinction confirms that unlike ReLU which loses capacity, and Rational which generates ineffective dimensions, ChannelWavAct successfully maintains a meaningful and rich representation space, allowing the backbone to continuously assimilate new features without plasticity loss.

\subsubsection{Generalizability}

Our preceding experiments have substantiated the efficacy of ChannelWavAct in maintaining the adaptability required for incoming data distributions. However, beyond mere adaptation, ensuring robust generalization to unseen data remains a paramount objective. We posit that generalization capability constitutes a critical dimension closely associated with the challenge of plasticity loss. To rigorously evaluate this attribute, we conduct assessments across two complementary continual learning paradigms: replay-based strategies and replay-free approaches.

\textbf{Replay-based approach} In the replay-based evaluation, we utilize the SCR framework \cite{mai2021supervised}, adopting the SSD method as the continual learning method \cite{gu2024summarizing}. This paradigm addresses catastrophic forgetting by maintaining a limited memory buffer of historical samples, which are interleaved with the current data stream during training. Utilizing a Reduced ResNet-18 backbone, our primary objective is to assess whether our proposed activation can significantly improve the model's plasticity in acquiring new tasks. To establish a rigorous benchmark, we compare our ChannelWavAct against a comprehensive spectrum of baselines, explicitly evaluating performance based on Last Accuracy (Last), Average Accuracy (Avg), and Forgetting (Fgt).

The generalization performance of different activation functions within the replay-based SSD framework is summarized in \Cref{tab:ssd_results}. As observed, ChannelWavAct consistently achieves superior performance across all three benchmark datasets, including Mini-ImageNet, CIFAR100, and Tiny-ImageNet. It secures the highest values in both Last and Average Accuracy metrics. In terms of Average Accuracy, ChannelWavAct achieves a gain of $2.0\%$ on Mini-ImageNet ($39.0\%$ vs. $37.0\%$), $3.2\%$ on CIFAR100 ($43.5\%$ vs. $40.3\%$) and $2.7\%$ on the challenging Tiny-ImageNet ($19.9\%$ vs. $17.2\%$). Moreover, even with a primary focus on enhancing plasticity, our method still demonstrates a comparable capability in mitigating catastrophic forgetting. These consistent improvements validate its exceptional capability to generalize effectively in replay-based continual learning scenarios.

\textbf{Replay-free approach} In the replay-free evaluation, we operate within the PyCIL framework \cite{zhou2023pycil}, implementing the EWC method \cite{kirkpatrick2017overcoming} and WA method \cite{zhao2020maintaining} using a standard ResNet-18 architecture. Unlike replay-based strategies, these approaches rely solely on parameter regularization or weight alignment to preserve past knowledge, thereby imposing a more rigorous test on the network's learning capacity. Consequently, our objective is to verify whether ChannelWavAct can maintain sufficient plasticity for new tasks even when model parameters are subject to regularization constraints. We benchmark our approach against the same set of baseline activations, using Average Accuracy (Avg) and Forgetting (Fgt) as key metrics \cite{zhou2023pycil} to quantify the trade-off between stability and plasticity.

The generalization performance of different activation functions within the replay-free framework is summarized in \Cref{tab:framework_comparison}. To more clearly isolate and compare the direct impact of activation designs on model performance, we first evaluate these functions in the absence of $L_2$ regularization. Furthermore, to assess the compatibility of different activation functions with standard plasticity-preserving techniques, we also provide experimental results under the inclusion of $L_2$ regularization. We exclude KAN on ImageNet100 due to severe parameter explosion. In the EWC setting, ChannelWavAct achieves $41.2\%$ average accuracy on CIFAR-100, outperforming the ReLU baseline by $7.8\%$. This advantage is even more pronounced in the WA framework, where ChannelWavAct exceeds ReLU by $22.6\%$ (CIFAR-100) and $28.7\%$ (ImageNet-100). These results demonstrate that our learnable wavelet structure provides the critical plasticity needed to acquire new knowledge without relying on memory buffers. The results under $L_2$ regularization further validate the robustness and compatibility of our method. While $L_2$ regularization typically restricts the update of learnable parameters, ChannelWavAct consistently maintains its leading position or stays within the top tier. For instance, in the WA setting on CIFAR-100, our method reaches $65.7\%$ average accuracy, proving its seamless integration with standard stability-preserving techniques. This versatility ensures that ChannelWavAct can be effectively deployed across diverse experimental configurations without compromising its intrinsic performance gains. The consistent performance of ChannelWavAct across both CIFAR and ImageNet-100 highlights its universality. Unlike other learnable activations that may suffer from parameter explosion or optimization instability on large-scale datasets, our method offers a scalable and stable solution. The synergy between the decoupled learning rate and the wavelet basis enables the model to achieve superior average accuracy while maintaining a competitive balance between plasticity and stability.
\begin{figure}[!t]
  \centering
  
  \subfloat[spectral learning dynamic curves.]{
    \includegraphics[width=0.45\textwidth]{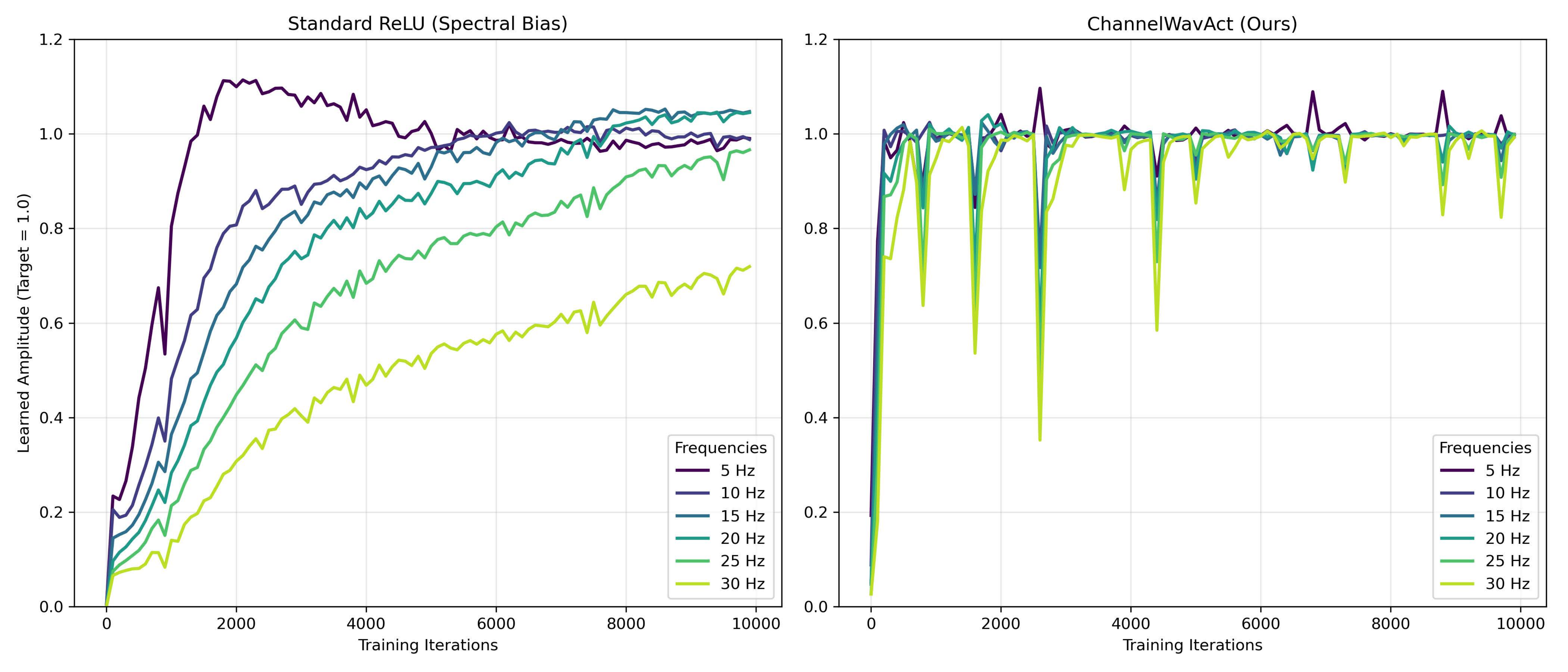}
    \label{fig:spectral_learning_dynamics}
  }
  
  \vspace{0.1in} 
  
  \subfloat[spectral learning heatmaps.]{
    \includegraphics[width=0.45\textwidth]{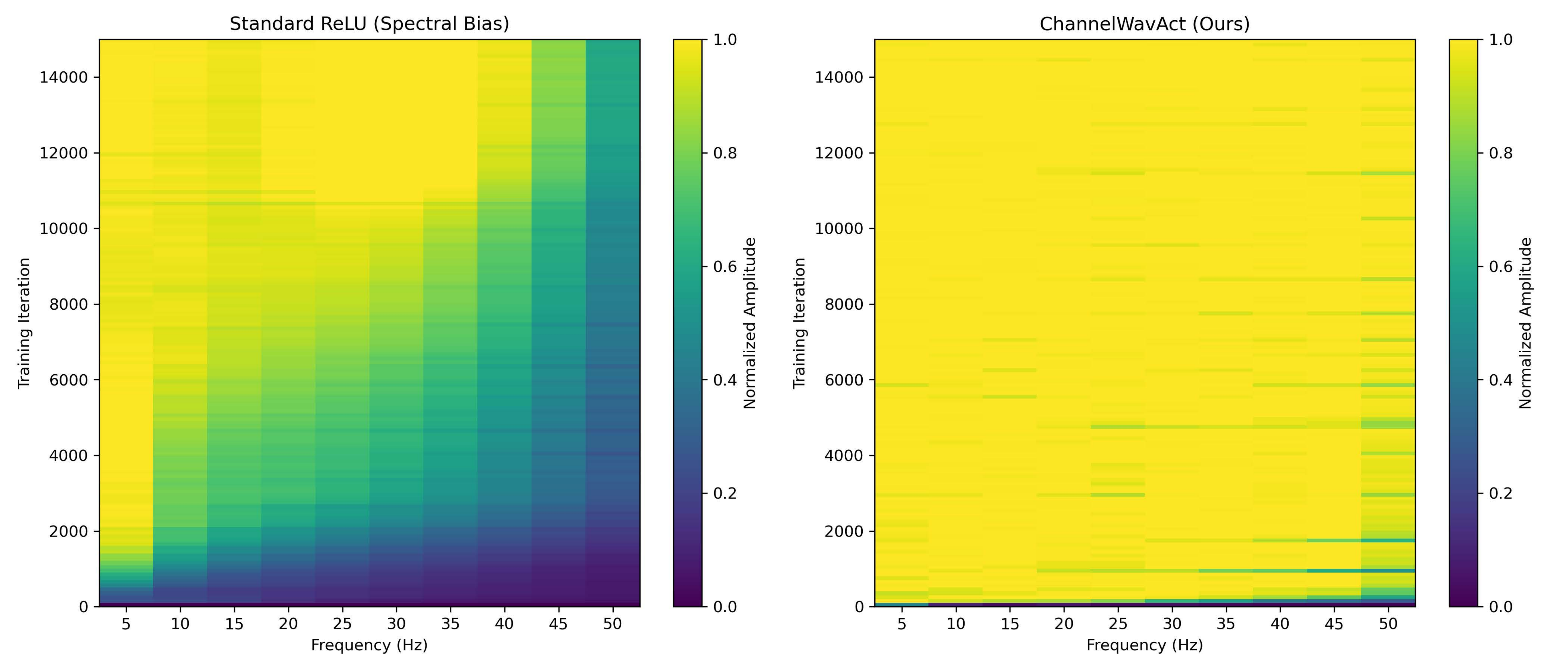}
    \label{fig:spectral_learning_heatmap}
  }
  
  \caption{
    (a) Evolution of the learned amplitude for frequency components (5 Hz to 30 Hz) during the training of a 1D synthetic regression task. Results are shown for the ReLU network (left) and ChannelWavAct (right), with a target amplitude of $1.0$. 
    (b) Heatmaps of the normalized amplitude across the frequency spectrum over training iterations for ReLU (left) and ChannelWavAct (right).
  }
  \label{fig:spectral_learning} 
\end{figure}

\begin{figure}[!t]
\centering
\includegraphics[width=\columnwidth]{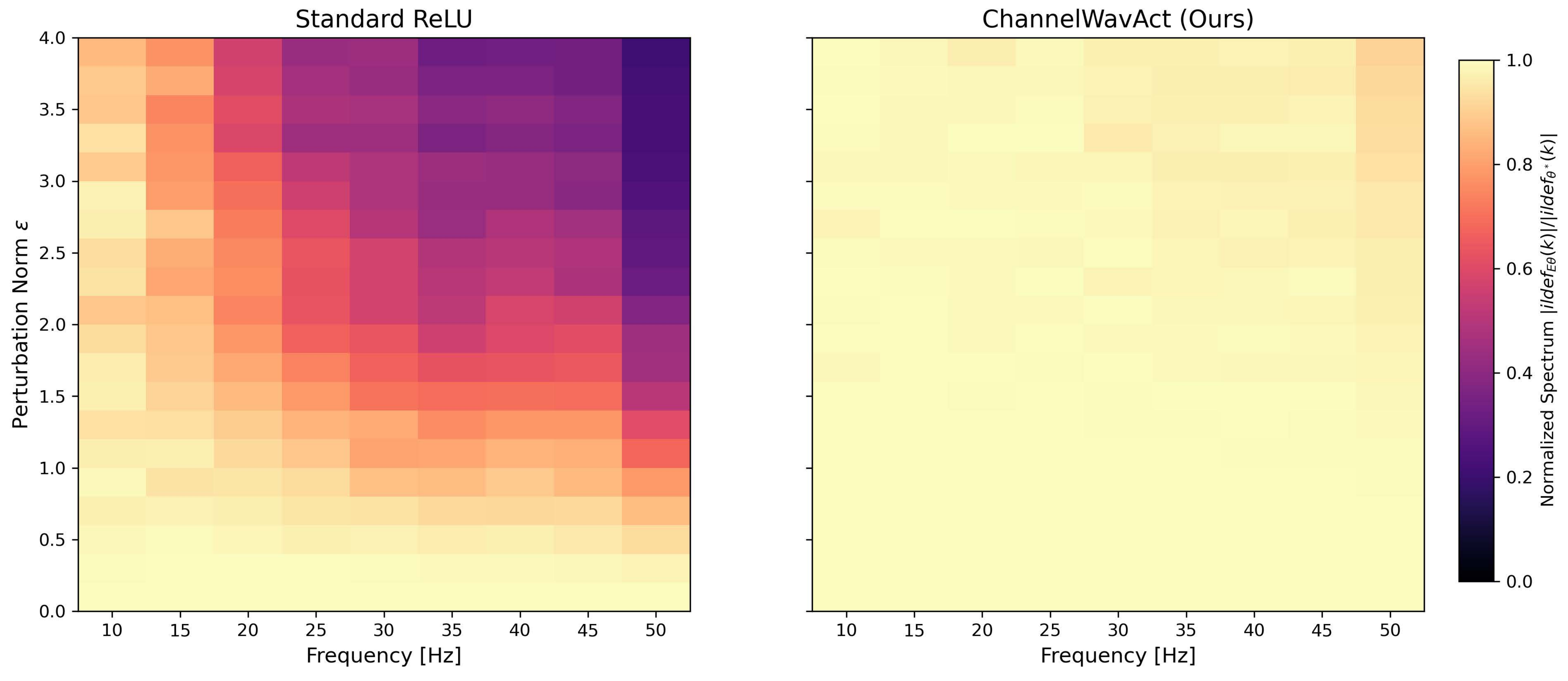}
\caption{Heatmaps of normalized spectral amplitude across frequencies ($x$-axis) and random parameter perturbation norms $\epsilon$ ($y$-axis) for the ReLU network (left) and ChannelWavAct (right).}
\label{fig:pertubation_robustness}
\end{figure}

\subsection{Spectral Analysis}

To demonstrate the effectiveness of ChannelWavAct in mitigating spectral bias, we conduct a series of spectral analysis experiments following the framework established by \cite{rahaman2019spectral}. These experiments compare the performance of standard ReLU networks with our proposed wavelet-based activation.

\subsubsection{Experiment 1}

We first examine the learning dynamics of different frequency components using a 1D regression task. The target signal is formulated as a superposition of six sinusoids with frequencies evenly spaced from 5 Hz to 30 Hz, all maintaining an equal target amplitude of 1.0. We train a four-layer Multi-Layer Perceptron (MLP) with a hidden dimension of 256 for 10,000 iterations, extracting the Fourier amplitude of each frequency component at regular intervals. 

As illustrated in \cref{fig:spectral_learning}, standard ReLU networks demonstrate a pronounced spectral bias. The lower frequency components converge to the target amplitude rapidly, whereas higher frequencies learn at a significantly slower rate and fail to fully reach the target amplitude even at the end of training. In contrast, ChannelWavAct utilizes its learnable Mexican Hat wavelet components to explicitly target these high-frequency variations. By injecting high-frequency wavelet bases, the model compensates for the inherent low-pass nature of standard activations, achieving an almost instantaneous convergence to the target amplitude across the entire frequency spectrum. Furthermore, while the dynamic adaptation process introduces brief, sharp perturbations during training, the network demonstrates remarkable structural robustness by recovering immediately, maintaining a uniform and accelerated learning capacity for all features regardless of their frequency.
\begin{figure}[!t]
  \centering
  
  \subfloat[fixed $k$.]{
    \includegraphics[width=0.45\textwidth]{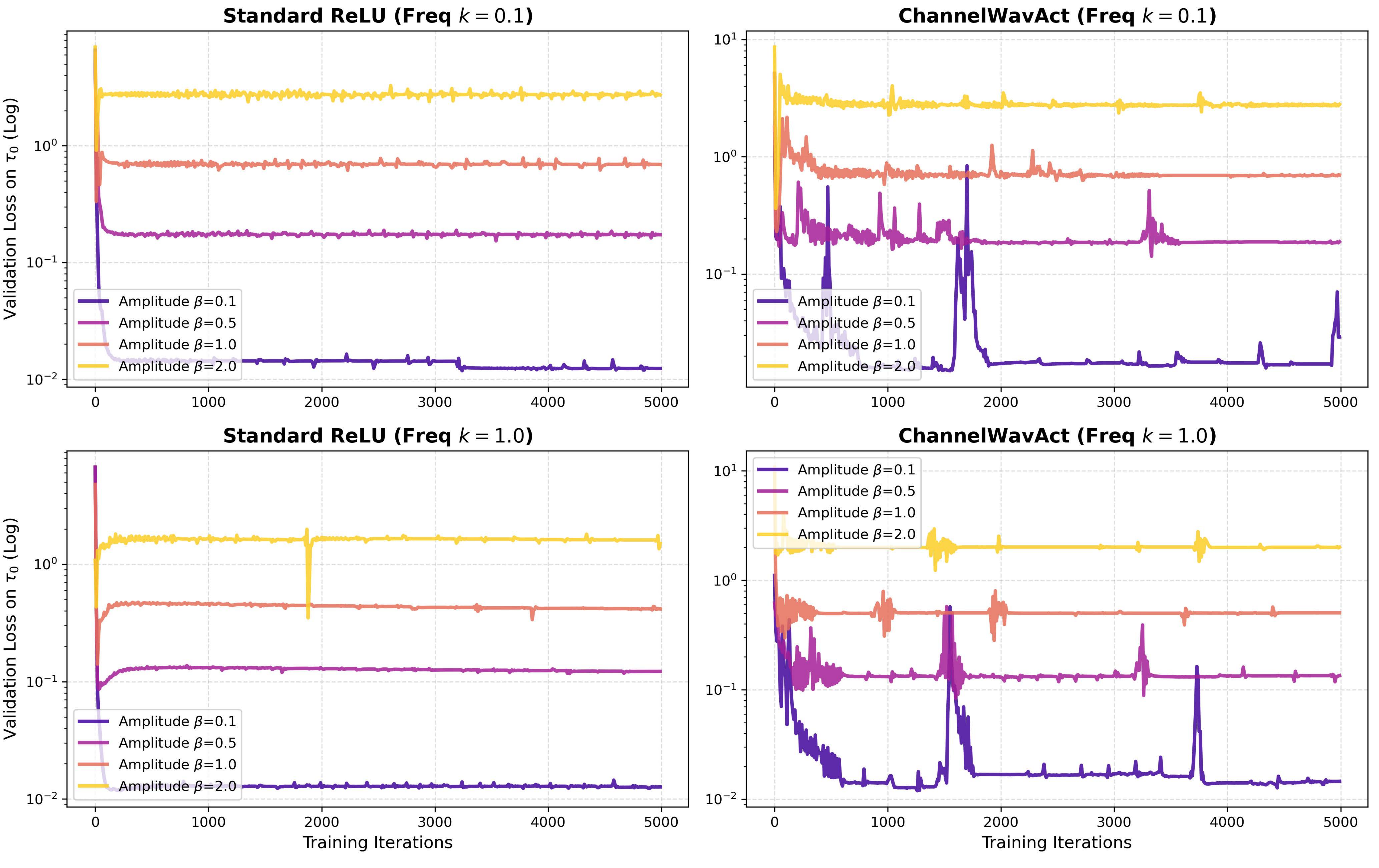}
    \label{fig:vary_beta}
  }
  
  \vspace{0.1in} 
  
  \subfloat[fixed $\beta$.]{
    \includegraphics[width=0.45\textwidth]{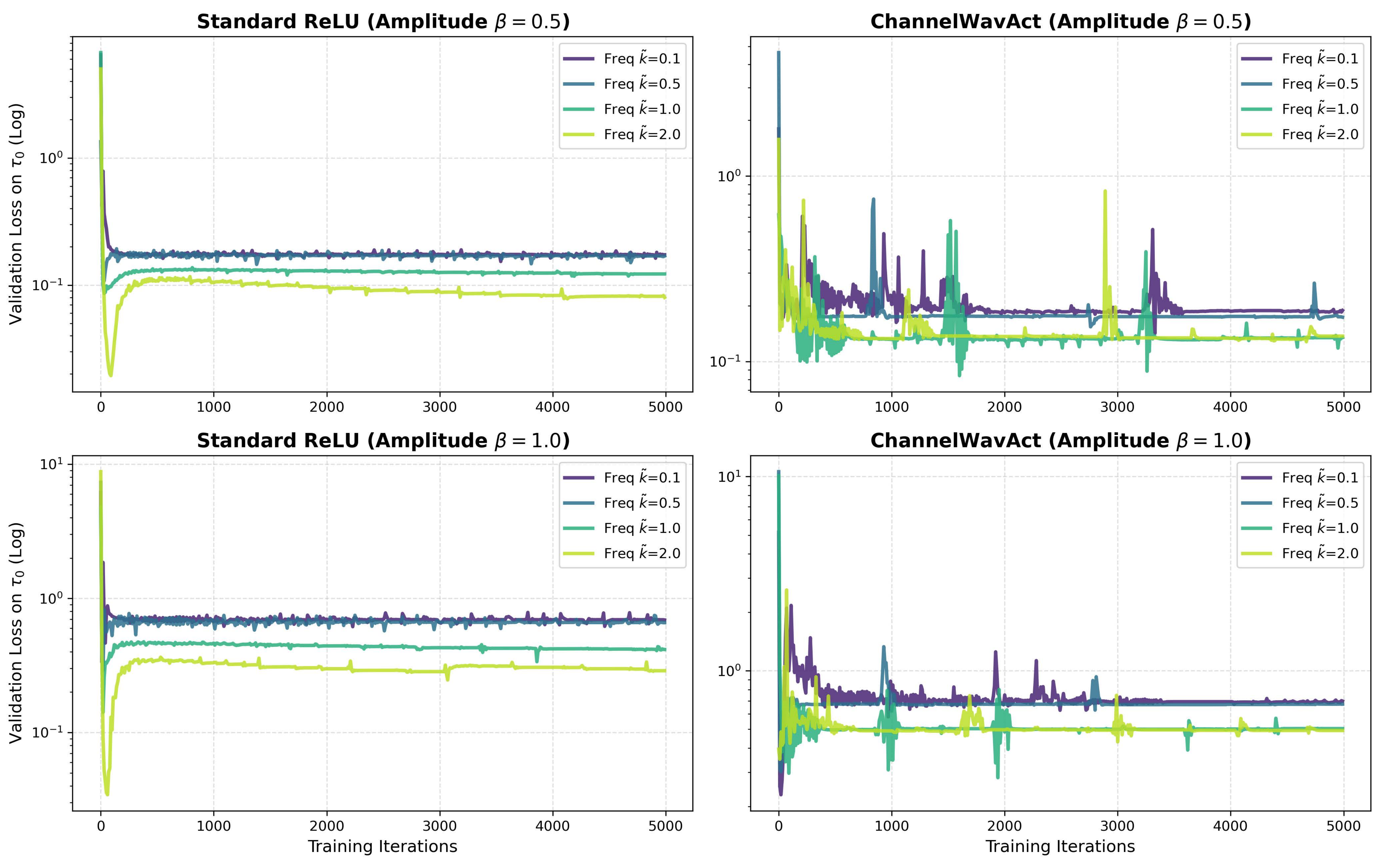}
    \label{fig:vary_k}
  }
  
  \caption{
    Validation loss on the clean target $\tau_0$ for ReLU (left) and ChannelWavAct (right), evaluated as a function of (a) noise amplitude $\beta$ with fixed frequencies $\tilde{k} \in \{0.1, 1.0\}$, and (b) spatial noise frequency $\tilde{k}$ with fixed amplitudes $\beta \in \{0.5, 1.0\}$.
  }
  \label{fig:mnist_vary_noise} 
\end{figure}

\subsubsection{Experiment 2}

To assess the structural resilience of learned frequency components against parameter drift, we evaluate the model's robustness to random parameter perturbations. Building upon the setup of Experiment 1, we apply isotropic random noise to the network parameters with varying perturbation norms ($\epsilon \in [0, 4.0]$) and average the resulting spectral magnitudes over 50 independent trials. 

As visually confirmed by \cref{fig:pertubation_robustness}, standard ReLU networks reveal a severe vulnerability under global parameter perturbations. While their low-frequency components exhibit relative stability, the normalized spectral amplitude of high-frequency components collapses rapidly even under minor perturbations. This demonstrates that high-frequency representations in fixed-activation networks occupy a highly sensitive and narrow volume in the parameter space. In contrast, ChannelWavAct maintains near-perfect spectral integrity across the entire frequency spectrum, even at the maximum perturbation norm. By explicitly decoupling high-frequency details into localized, multi-scale wavelet bases, ChannelWavAct structurally anchors complex features. This structural independence ensures that subsequent parameter fluctuations during the acquisition of new tasks do not catastrophically erase previously consolidated high-frequency knowledge.
\begin{figure}[!t]
\centering
\includegraphics[width=\columnwidth]{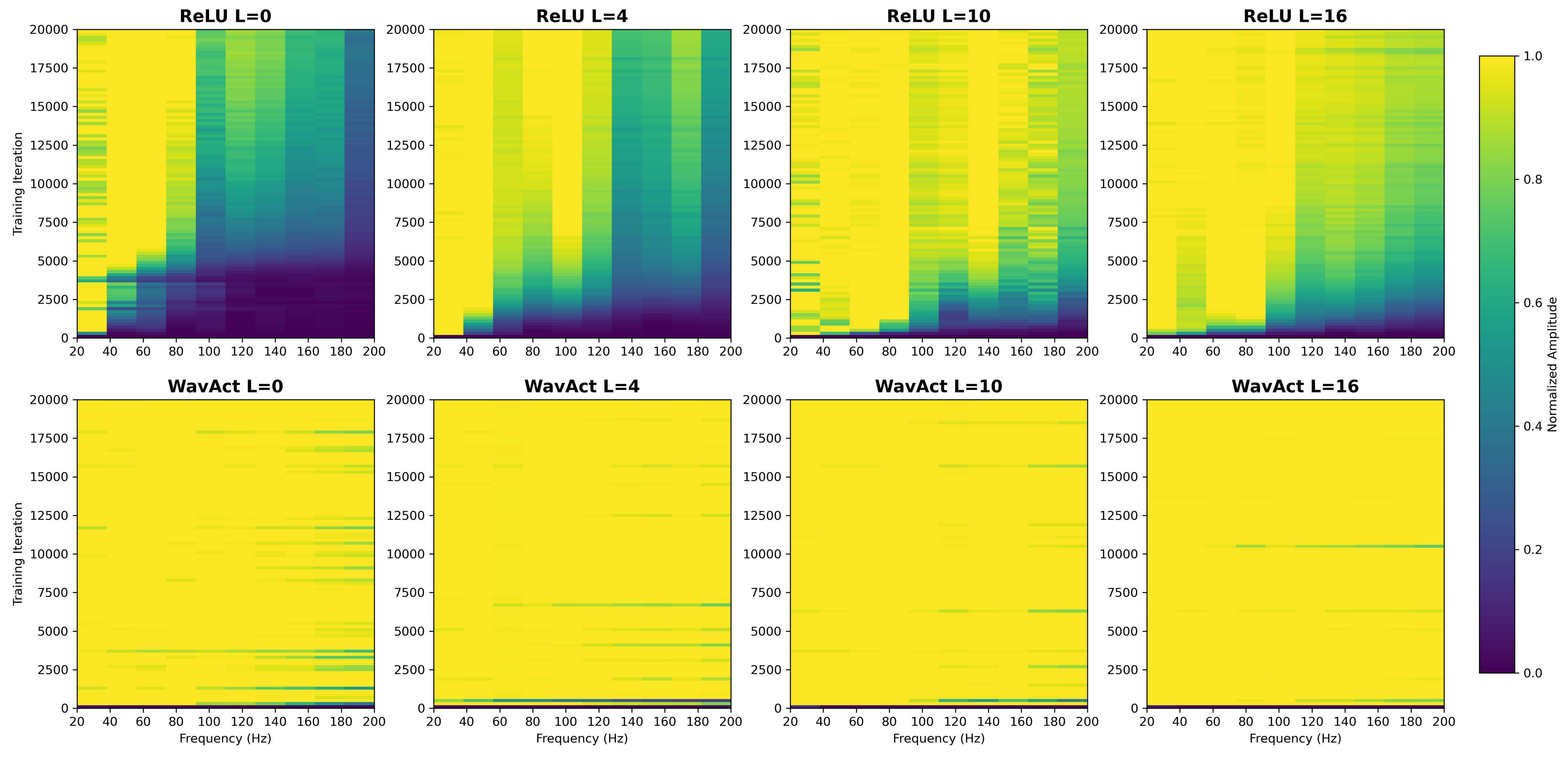}
\caption{Heatmaps of the normalized spectral amplitude across learned frequencies ($x$-axis) and 20,000 training iterations ($y$-axis) for target signals defined on 2D flower-shaped manifolds $\gamma_L$ with $L \in \{0, 4, 10, 16\}$. The visualization shows results for standard ReLU networks (top) and ChannelWavAct (bottom).}
\label{fig:manifold_heatmap}
\end{figure}

\subsubsection{Experiment 3}

We extend our analysis to high-dimensional data by training a Multi-Layer Perceptron (MLP) on a binary classification subset of MNIST (digits 3 and 8). To test the model's ability to distinguish between relevant semantic features and frequency-dependent noise, we construct a noisy target $\tau_k = \tau_0 + \beta \sin(\tilde{k}\|x\|)$, where $\tau_0$ is the clean label, $\beta$ is the noise amplitude, and $\tilde{k}$ is the spatial frequency of the radial noise. We evaluate the validation loss on the clean target $\tau_0$ across varying amplitudes $\beta \in \{0.1, 0.5, 1.0, 2.0\}$ and frequencies $\tilde{k} \in \{0.1, 0.5, 1.0, 2.0\}$.

The experimental results in \cref{fig:mnist_vary_noise} validate that standard ReLU networks exhibit a strong spectral bias, leaving them highly susceptible to low-frequency noise. As observed in the standard ReLU validation curves, low-frequency noise $\tilde{k}=0.1$ is rapidly and preferentially fitted by the network, resulting in a significantly elevated and persistent validation loss that worsens as the noise amplitude $\beta$ increases. Furthermore, when exposed to high-frequency noise $\tilde{k} \geq 1.0$, ReLU networks demonstrate a distinct dip in validation loss early in training before overfitting the noise, confirming their delayed capability to represent high-frequency variations. Conversely, ChannelWavAct leverages its multi-resolution wavelet bases to actively mitigate this bias. The validation loss curves for ChannelWavAct display profound resilience: the network achieves a substantially lower steady-state validation loss across all noise amplitudes and frequencies.

\subsubsection{Experiment 4}

Finally, we investigate the interaction between data manifold geometry and the network's frequency learning capabilities. We construct a 2D flower-shaped manifold $\gamma_L(z)$ parameterized by a latent variable $z \in [0, 1]$, where the parameter $L \in \{0, 4, 10, 16\}$ controls the number of petals and the overall topological curvature. The network, a six-layer Multi-Layer Perceptron, is tasked with regressing a highly complex target signal comprising ten superimposed sinusoids with frequencies ranging from $20$ Hz to $200$ Hz defined on the latent space. The training iterations is set to be $20000$.

The results in \cref{fig:manifold_heatmap} shows that standard ReLU networks are strictly constrained by the manifold's geometric properties. On a simple circular manifold $L=0$, ReLU completely fails to capture high-frequency components. As the manifold complexity increases to $L=16$, the network's ability to fit higher frequencies gradually improves, aligning with the theoretical premise that complex topologies allow standard networks to represent high latent frequencies using lower ambient frequencies. However, ReLU still exhibits significantly delayed convergence and struggles to fully resolve the 180–200 Hz band. Conversely, ChannelWavAct effectively decouples the model's representational capacity from the underlying data geometry. By dynamically injecting multi-scale wavelet bases, it possesses an intrinsic capability to resolve high-frequency details regardless of the manifold's spatial curvature. The experimental heatmaps demonstrate that ChannelWavAct achieves near-instantaneous and uniform convergence across the entire 20–200 Hz spectrum for all values of $L$. This structural independence from manifold complexity ensures that the model can rapidly assimilate intricate target functions across arbitrary topological structures.
\begin{table}[!t]
\caption{Ablation study on the effectiveness of each component in ChannelWavAct on the CIFAR100 dataset.\label{tab:ablation_on_modules}}
\centering
\setlength{\tabcolsep}{6pt}
\begin{tabular}{|l|ccccccc|}
\hline
Module & ReLU & Wav. & Inj. & Reg. & Last & Avg & Fgt \\
\hline
$\#1$ & $\checkmark$ & & & & 28.3 & 40.3 & 18.5 \\
$\#2$ & & $\checkmark$ & & & 29.0 & 42.5 & 17.1 \\
$\#3$ & & $\checkmark$ & $\checkmark$ & & 29.6 & 43.3 & 17.7 \\
$\#4$ & & $\checkmark$ & & $\checkmark$ & 28.8 & 42.2 & 16.5 \\
$\#5$ & & $\checkmark$ & $\checkmark$ & $\checkmark$ & \textbf{30.2} & \textbf{43.5} & \textbf{16.4} \\
\hline
\end{tabular}
\end{table}

\begin{table}[!t]
\caption{Ablation study on the impact of decoupled learning rates on the CIFAR100 dataset.\label{tab:ablation_lr}}
\centering
\setlength{\tabcolsep}{6pt}
\begin{tabular}{|l|ccccc|}
\hline
Module & ReLU & ChaWav. & Decoupled. & Avg & Fgt \\
\hline
$\#1$ & $\checkmark$ & & & 33.4 & 56.4 \\
$\#2$ & $\checkmark$ & & $\checkmark$ & 38.9 & 43.7 \\ 
$\#3$ & & $\checkmark$ & & 36.5 & 52.3 \\
$\#4$ & & $\checkmark$ & $\checkmark$ & 41.2 & 45.3 \\
\hline
\end{tabular}
\end{table}

\subsection{Ablation}

\subsubsection{Component analysis}

To validate the efficacy of our proposed method, we conducted comprehensive ablation studies on the CIFAR-100 dataset. The results in \cref{tab:ablation_on_modules} demonstrate that replacing fixed ReLU with our learnable ChannelWavAct yields immediate accuracy gains ($42.5\%$ vs. $40.3\%$), confirming the superiority of learnable wavelet bases. While dynamic wavelet injection further enhances plasticity ($43.3\%$), optimal performance is achieved only when synergized with stability regularization, which effectively minimizes forgetting to $16.4\%$ and resolves the stability-plasticity dilemma. Additionally, we validate the criticality of our decoupled optimization strategy in \cref{tab:ablation_lr}. By categorizing parameters into plasticity and stability groups, it improves average accuracy by $4.7\%$ and reduces forgetting by $7.0\%$. These results confirm that differential optimization facilitates rapid restructuring to prevent under-fitting while safeguarding against uncontrolled representation drift during adaptation.
\begin{figure}[!t]
  \centering

  \subfloat[$\Delta$.]{
    \includegraphics[width=0.45\textwidth]{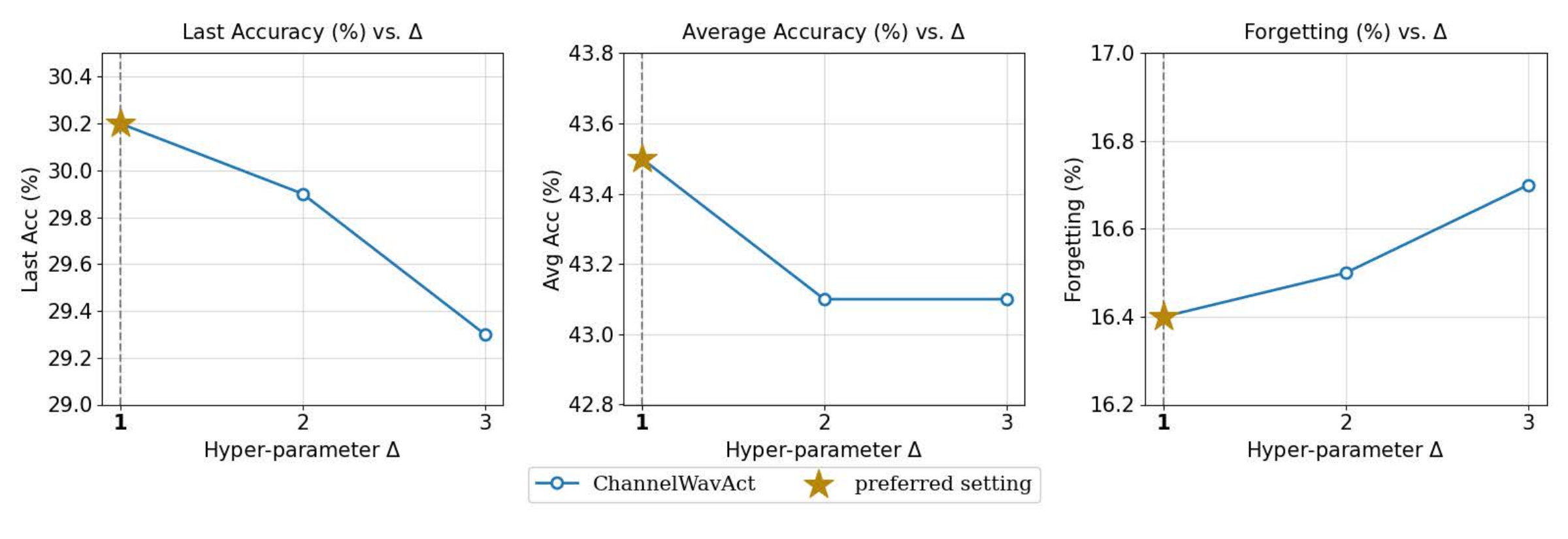}
    \label{fig:inject number Delta}
  }
  
  \subfloat[$P$.]{
    \includegraphics[width=0.45\textwidth]{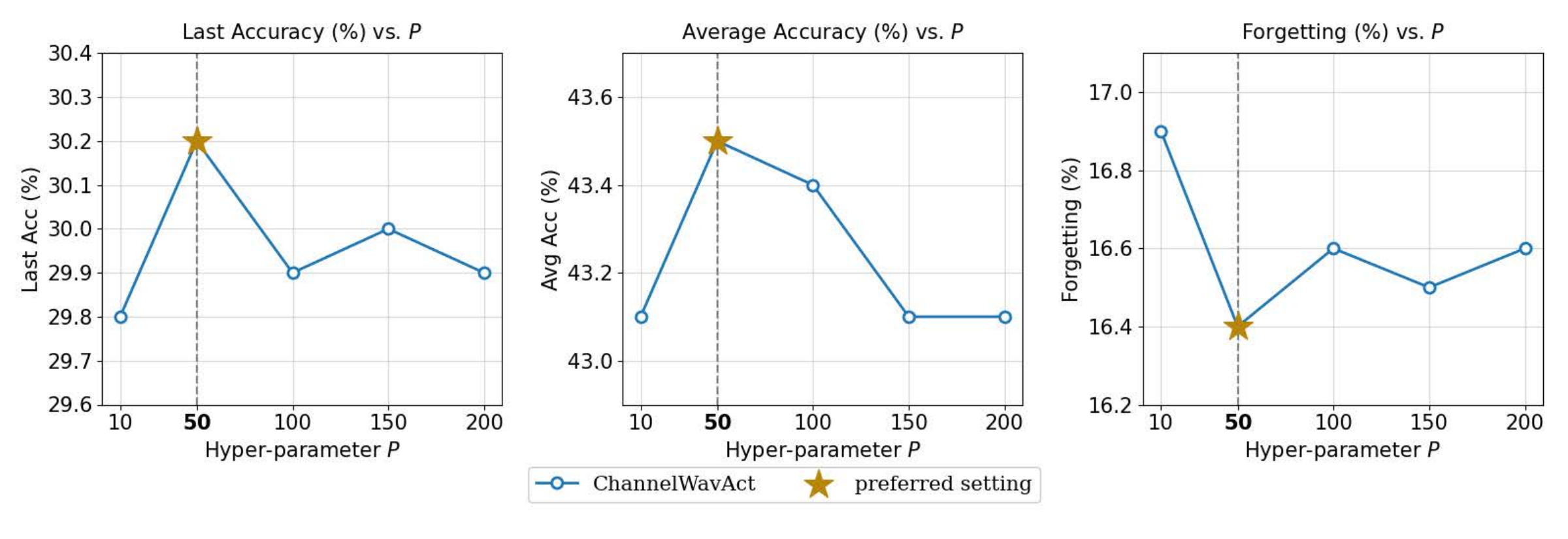}
    \label{fig:patience threshold P}
  }
  
  \subfloat[$\lambda$.]{
    \includegraphics[width=0.45\textwidth]{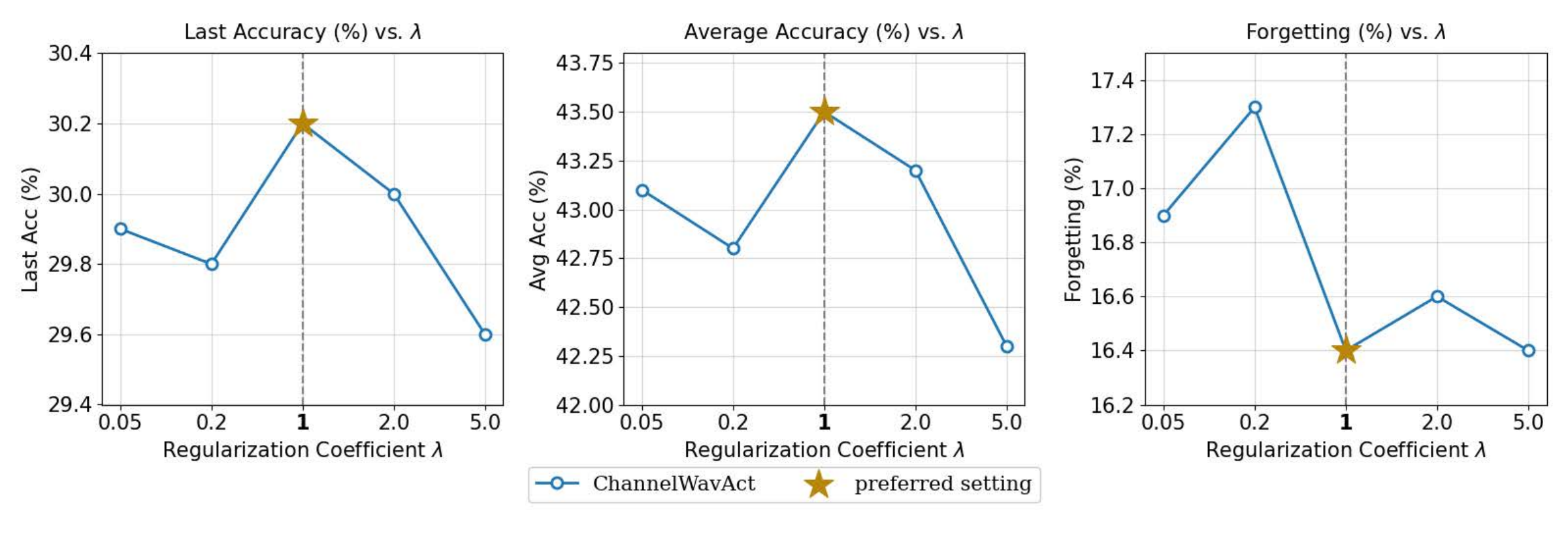}
    \label{fig:reg lambda}
  }
  
  \subfloat[$\delta$.]{
    \includegraphics[width=0.45\textwidth]{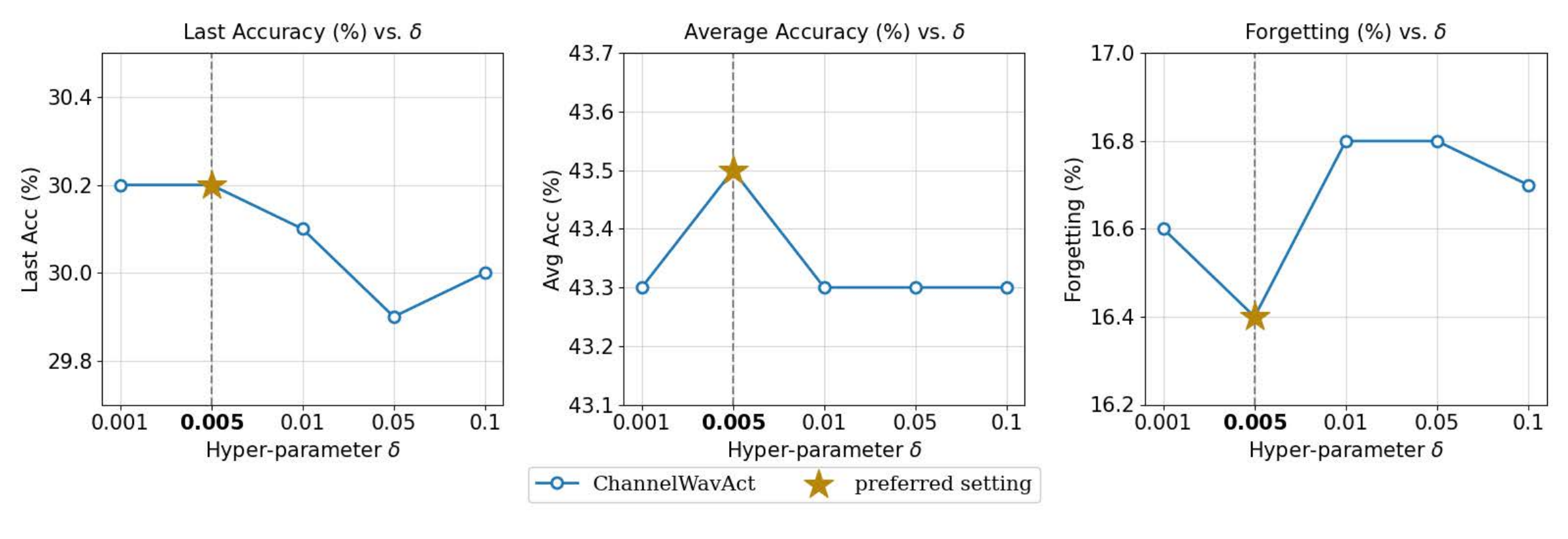}
    \label{fig:relative margin delta}
  }
  
  \caption{
    Impact of key hyperparameters on continual learning performance, evaluated across: (a) the number of injected wavelet bases $\Delta$ per expansion phase, (b) the patience threshold $P$ for injection triggering, (c) the regularization coefficient $\lambda$ applied to pre-existing wavelet weights and (d) the relative margin $\delta$ used for loss stagnation detection.
  }
  \label{fig:parameters_sensitivity} 
\end{figure}

\subsubsection{Parameter Sensitivity Analysis}

To thoroughly evaluate the robustness of ChannelWavAct, we conduct a parameter sensitivity analysis on its core hyperparameters: the injection number $\Delta$, patience threshold $P$, regularization coefficient $\lambda$, and relative margin $\delta$ in \cref{fig:parameters_sensitivity}. The impact of these parameters is measured across three key continual learning metrics: Last Accuracy, Average Accuracy, and Forgetting.

\textbf{Injection Number $\Delta$:} The hyperparameter $\Delta$ controls the number of new wavelet bases injected during each capacity expansion phase. As illustrated in \cref{fig:inject number Delta}, the model achieves its peak performance at a minimal injection size of $\Delta=1$. Increasing the injection number leads to a noticeable drop in both Average and Last Accuracy, accompanied by an increase in forgetting. This suggests that injecting too many new parameters simultaneously disrupts the optimization landscape and may lead to redundant capacity, whereas a conservative, incremental expansion provides sufficient representational power while maintaining network stability.

\textbf{Patience Threshold $P$:} The patience threshold determines how many iterations the model waits during loss stagnation before triggering a dynamic injection. The analysis in \cref{fig:patience threshold P} reveals an optimal balance at $P=50$. Setting the threshold too low degrades performance, likely because it triggers premature injections before the model has fully converged using its existing capacity. Conversely, a threshold that is too high delays necessary expansions, forcing the model to struggle with underfitting on complex new data distributions. $P=50$ effectively ensures that capacity is expanded only when learning has genuinely plateaued.

\textbf{Regularization Coefficient $\lambda$:} The coefficient $\lambda$ dictates the penalty strength applied to pre-existing wavelet weights to prevent catastrophic forgetting. The results in \cref{fig:reg lambda} show a distinct performance peak at $\lambda=1$. When the regularization is too weak, the network suffers from severe forgetting, as it rapidly overwrites older task representations to accommodate new ones. On the other hand, excessive regularization overly constrains the parameter space, causing a steep drop in accuracy due to a loss of plasticity. Setting $\lambda=1$ optimally navigates the stability-plasticity dilemma.

\textbf{Relative Margin $\delta$:} The margin $\delta$ defines the relative loss improvement required to reset the patience counter. The model demonstrates optimal performance at a tight margin of $\delta=0.005$. A larger margin makes the stagnation detector overly insensitive, preventing the network from injecting capacity when it is actually needed. Conversely, an extremely small margin slightly increases forgetting, as the mechanism may overreact to standard mini-batch noise rather than true convergence. $\delta=0.005$ provides the most reliable signal for actual learning plateaus.

\section{Conclusion}

In this paper, we address the challenge of plasticity loss in continual learning by proposing \textbf{ChannelWavAct}, which decomposes the activation function into a global base approximation and local wavelet details. Through dynamic wavelet injection and stability regularization, our approach effectively harmonizes the stability-plasticity trade-off. Crucially, we provide a rigorous theoretical foundation that validates the necessity of the hybrid activation structure and the decoupled optimization strategy in overcoming spectral bias. We further establish the mathematical necessity of the loss-based trigger mechanism and demonstrate the superior $L^2$ approximation capability of ChannelWavAct for complex target signals. Extensive experiments confirm that our method achieves state-of-the-art performance with superior trainability and generalization across diverse benchmarks. 

While the current implementation incurs a relatively higher computational overhead, future work will focus on enhancing the computational efficiency of learnable wavelets and establishing more comprehensive theoretical frameworks regarding optimization and generalization dynamics.


\end{document}